\documentclass{article}

\usepackage{PRIMEarxiv}
\usepackage{makecell}
\usepackage[utf8]{inputenc} % allow utf-8 input
\usepackage[T1]{fontenc}    % use 8-bit T1 fonts
\usepackage{hyperref}       % hyperlinks
\usepackage{url}            % simple URL typesetting
\usepackage{booktabs}       % professional-quality tables
\usepackage{amsfonts}       % blackboard math symbols
\usepackage{nicefrac}       % compact symbols for 1/2, etc.
\usepackage{microtype}      % microtypography
\usepackage{lipsum}
\usepackage{fancyhdr}       % header
\usepackage{graphicx}  
\usepackage{colortbl}
\usepackage{graphicx}
\usepackage{listings}
\usepackage{manyfoot}
 \usepackage{siunitx} 
\usepackage{mathrsfs}
\usepackage[utf8]{inputenc}
\usepackage{multirow}
\usepackage{placeins}
\usepackage{ragged2e}
\usepackage{subcaption}
\usepackage{tabularx}
\usepackage{textcomp}
\usepackage{threeparttable}
\usepackage{tikz}
\usepackage{url}
\usepackage[table]{xcolor}% graphics
\graphicspath{{media/}}     % organize your images and other figures under media/ folder

\title{Why AI-Generated Text Detection Fails: Evidence from Explainable AI Beyond Benchmark Accuracy
}

\author{
  Shushanta Pudasaini \\
  Technological University Dublin \\
  \texttt{D23129142@mytudublin.ie} \\
   \And
  Luis Miralles-Pechuán\\
  Technological University Dublin\\
  \texttt{luis.miralles@TUDublin.ie} \\
  \AND
  David Lillis \\
  University College Dublin \\
  \texttt{david.lillis@ucd.ie} \\
  \And
  Marisa Llorens Salvador \\
  Technological University Dublin\\
  \texttt{marisa.llorens@TUDublin.ie} \\
}

\begin{document}
\maketitle

\begin{abstract}

% As Large Language Models (LLMs) become widely used in education and online communication, distinguishing AI-generated text from human writing has emerged as a critical challenge. While many detection systems report high benchmark accuracy, their real-world reliability remains uncertain, particularly in high-stakes settings such as academic assessment.

% The widespread adoption of Large Language Models (LLMs) has made AI-Generated Text detection a pressing and complex problem. While many detection systems report high benchmark accuracy, their real-world reliability remains uncertain, and interpretability of these systems is not explored.

% This paper investigates whether modern detectors truly identify machine authorship or instead learn dataset-specific artefacts. We introduce an interpretable detection framework that combines linguistic feature engineering, machine learning, and explainable AI. Across two major benchmark corpora (PAN-CLEF 2025 and COLING 2025), models trained on 38 linguistic features achieve leaderboard-competitive performance, reaching an F1 score of 0.9734 without relying on huge LLMs.

The widespread adoption of Large Language Models (LLMs) has made the detection of AI-generated text a pressing and complex challenge. Although many detection systems report high benchmark accuracy, their reliability in real-world settings remains uncertain, and their interpretability is often unexplored. In this work, we investigate whether contemporary detectors genuinely identify machine authorship or merely exploit dataset-specific artefacts. We propose an interpretable detection framework that integrates linguistic feature engineering, machine learning, and explainable AI techniques. Evaluated across two major benchmark corpora—PAN-CLEF 2025 and COLING 2025—models trained on 30 optimal linguistic features achieve leaderboard-competitive performance, attaining an F1 score of 0.9734 on the PAN-CLEF dataset and an F1 score of 0.8025 on the COLING dataset.

 However, systematic cross-domain and cross-generator evaluation reveals substantial generalisation failure: classifiers that excel in-domain degrade significantly under distribution shift. Using SHAP-based explanations, we show that the most influential features differ markedly between datasets, indicating that detectors often rely on dataset-specific stylistic cues rather than stable signals of machine authorship. Further investigating this, we perform an in-depth error analysis, applying SHAP to False Negatives and False Positives from the model. 
 % Ensemble models partially improve robustness but do not eliminate this effect.

Our findings show that benchmark accuracy is not reliable evidence of authorship detection: strong in-domain performance can coincide with substantial failure under domain and generator shift. Our in-depth error analysis exposes a fundamental tension in linguistic-feature-based AI text detection: the features that are most discriminative on in-domain data are also the features most susceptible to domain shift, formatting variation, and text-length effects. We believe that this knowledge helps in building AI detectors that are robust under different settings.  To support replication and practical use, we release an open-source Python package that returns both predictions and instance-level explanations for individual texts.

% We argue that explainability should be treated not only as a transparency aid, but as a validity diagnostic that reveals what evidence a detector is actually using. In educational and other high-stakes contexts, detectors should therefore be used only as probabilistic, explainable support for human judgment and never as an automated basis for punitive decisions. To support replication and practical use, we release an open-source Python package that returns both predictions and instance-level explanations for individual texts.
\end{abstract}

\begin{center}
  \vspace{-1em}
  \href{https://shushantatud.github.io/ExplainAIGeneratedText/}{\textbf{Project Page: https://shushantatud.github.io/ExplainAIGeneratedText/}}
\end{center}

% keywords can be removed

\section{Introduction}
The rapid adoption of Large Language Models (LLMs) has made the distinction between human-written and AI-generated text a central concern in education, publishing, and online communication. Automated detectors are increasingly proposed as a solution, and recent shared tasks report very high accuracy on benchmark datasets. However, despite these results, the reliability of AI-generated text detection in real-world settings remains uncertain.

A particularly consequential failure mode is the False Positive (FP) — an instance in which a human-authored text is misclassified as AI-generated. For university students, the ramifications of such misclassification can be severe, ranging from academic penalties to reputational harm. Turnitin, one of the most widely deployed commercial AI detection platforms, reports a False Positive Rate (FPR) of below 4\% at the sentence level and below 1\% at the document level \cite{chechitelli2023falsepositive}. While these figures may appear reassuring in isolation, they become far more troubling at scale: even a sub-1\% error rate, applied across millions of student submissions, translates into a substantial number of wrongful flags. 

Similarly, False Negatives (FN) - an instance in which an AI-generated text bypasses the detector. As recent LLMs grow more sophisticated, their outputs increasingly approximate the stylistic and syntactic patterns of human writing, making detection even more difficult. This challenge is further compounded by a growing arsenal of evasion techniques — including paraphrasing, prompt engineering — which are specifically designed to fool the AI detectors. Thus, the true prevalence of undetected AI-generated content may be considerably higher than current benchmarks suggest.

In high-stakes contexts such as academic assessment, therefore, the central question is not whether detectors can achieve strong performance on controlled benchmarks, but whether they can reliably and fairly identify machine authorship in the diverse, open-ended conditions of real-world use.

% False Positives(FP)(text which are written by humans but flagged as AI-generated) can be a very big problem for university students. Turnitin (a widely used commercial AI detector) report that they have less than 4\% False Positive Rate (FPR) for sentence-level texts and less than 1\% FPR for documents. 

% In high-stakes contexts such as academic assessment, the critical question is not whether detectors can perform well on benchmarks, but whether they genuinely identify machine authorship.

A key unresolved issue is whether modern detectors learn universal characteristics of AI-generated language or instead exploit dataset-specific artefacts. Benchmark corpora differ substantially in domain, genre, writing conditions, and generator models. Consequently, high in-domain performance may reflect the ability to distinguish particular datasets rather than to detect AI authorship itself. If this is the case, reported accuracy may overestimate real-world capability, especially when detectors are applied to unseen domains or newly released language models.

From a measurement perspective, an AI-text detector is an instrument that claims to measure a latent construct: \emph{machine authorship}. High benchmark scores demonstrate only that a model separates the labels within a specific dataset; they do not, by themselves, establish that the detector is measuring authorship rather than confounded properties of dataset construction (e.g., topic, genre, prompt style, length, or collection pipeline). We therefore treat \emph{detector validity} as a first-class outcome: a detector is valid only if the evidence it uses remains stable under domain and generator shift and aligns with plausible linguistic signals of authorship rather than corpus-specific cues.

Educational settings illustrate the importance of this problem. Surveys indicate that a large proportion of students regularly use generative AI tools such as ChatGPT, Grammarly, and Microsoft Copilot \cite{freeman2025studentai}. While such tools can support learning, they also enable the generation of complete assignments with minimal human effort. Institutions, therefore, face increasing pressure to ensure fair assessment and academic integrity. At the same time, unreliable detection systems risk unjust decisions and reduced trust in automated evaluation.

Existing technical approaches to AI-generated text detection include watermarking-based methods, zero-shot detectors, and supervised classifiers \cite{ghosal2023towards}. These approaches are commonly evaluated in shared tasks such as SemEval 2024 \cite{wang-etal-2024-semeval-2024}, PAN@CLEF 2025 \cite{bevendorff:2025}, and COLING 2025 \cite{genaidetect-2025-genai}. However, their validity remains contested \cite{sadasivan2023can}. Watermarking methods are vulnerable to paraphrasing and theoretical limitations \cite{zhang2025watermarkssandimpossibilitystrong}, while many high-performing detectors rely on opaque large models that provide little justification for their decisions \cite{ji2025detectingmachinegeneratedtextsjust}. This lack of transparency is particularly problematic in high-stakes applications.

Linguistic-feature-based approaches offer an alternative by grounding predictions in interpretable properties of text. Prior works have identified stylometric and linguistic signals that distinguish human and machine writing within individual datasets \cite{opara2024styloai, tervcon2025linguistic}. 
% However, it remains unclear whether these signals generalise across domains and generators, and most studies report aggregate feature importance without explaining individual predictions. To address this gap, we combine linguistic features with machine learning algorithms and explainable AI techniques to investigate two central research questions. First, we ask which linguistic features most effectively distinguish AI-generated text from human-written text across diverse domains and generators. Second, we examine whether machine learning models trained on these features genuinely capture the characteristics of machine authorship or merely learn dataset-specific artefacts, evaluating this through systematic cross-domain and cross-generator testing.
However, it remains unclear whether the linguistic signals identified in prior works reflect genuine characteristics of machine authorship or are instead artefacts of the specific datasets and generators on which detectors are trained. Most existing studies report aggregate feature importance without explaining individual predictions, and few subject their models to systematic evaluation beyond the conditions of their training distribution. To address this gap, we combine linguistic features with machine learning algorithms and explainable AI techniques to investigate the following central research question: do machine learning models trained on linguistic features genuinely capture the underlying properties of machine authorship, or do they merely learn dataset-specific artefacts? We evaluate this through rigorous cross-domain and cross-generator testing.

This paper investigates the reliability of AI-generated text detection rather than only its accuracy. We test the hypothesis that many detectors rely on dataset-specific stylistic cues instead of stable indicators of machine authorship. To examine this, we introduce an interpretable detection framework combining linguistic feature engineering, classical machine learning, and explainable AI. We evaluate models across two major benchmark datasets (PAN CLEF and COLING) under in-domain, cross-domain, and cross-generator conditions, and analyse their behaviour using SHAP (SHapley Additive exPlanations).

Our central claim is that explainability can be used as a diagnostic tool. By analysing feature contributions for individual predictions, we identify the linguistic patterns actually used by the models and determine whether they represent generalisable authorship signals or dataset artefacts. This enables direct examination of why detectors succeed on benchmarks yet fail under domain and generator shift.

The primary contribution of this work is an empirical validity analysis of AI-generated text detection. We show that models can achieve near state-of-the-art benchmark performance while failing under cross-domain and cross-generator evaluation, and we use instance-level explanations to diagnose \emph{why} this happens. Specifically, we demonstrate that the features driving predictions differ markedly across datasets, indicating reliance on corpus-specific stylistic cues rather than stable signatures of machine authorship. As a secondary contribution, we provide an interpretable feature-based framework and an open-source implementation that returns both predictions and instance-level explanations to support reproducible evaluation and responsible use in academic integrity workflows. For reproducibility purposes, we made our code available on GitHub.

\section{Literature Review}

Research on AI-generated text detection has grown rapidly alongside the growing use of LLMs. Prior work broadly addresses four aspects of the problem: detection methodologies, linguistic feature-based approaches, robustness and generalisation, and explainability. Taken together, these strands reveal a central unresolved issue: although detectors often achieve high benchmark accuracy, it remains unclear whether they identify machine authorship itself or instead exploit dataset-specific characteristics.

\subsection{Approaches to AI-Generated Text Detection}

A wide range of techniques has been proposed for detecting AI-generated text, including statistical analysis of model outputs (e.g., perplexity, token probability distributions, compression metrics, and stylometric indicators) and deep neural approaches that rely on representations learned by large models \cite{yang2025imitation}. Surveys typically group these methods into watermarking-based approaches, zero-shot detectors, supervised classifiers, and adversarial strategies \cite{pudasaini2024survey, hu2023radar}.

Evaluation has largely been driven by shared tasks and benchmarks, including the Voight--Kampff Generative AI Detection task at PAN@CLEF 2025 \cite{bevendorff:2025} and the RAID benchmark \cite{dugan-etal-2024-raid}. In these competitions, baseline systems often rely on zero-shot detection (e.g., Binoculars, FastDetectGPT), while top-ranked submissions frequently fine-tune transformer models on task-specific datasets. Systems such as \emph{mdok of KInIT} \cite{macko:2025} and \emph{e5-small-lora-ai-generated-detector} \cite{dugan-etal-2024-raid} achieve strong leaderboard performance using this approach.

However, benchmark success does not necessarily imply reliable authorship detection. Many high-performing systems operate as black boxes, offering little justification for individual decisions and limited interpretability. Moreover, performance often degrades when models are applied to unseen domains or newer generators \cite{pudasaini-etal-2025-benchmarking}, raising the possibility that detectors distinguish datasets rather than human and machine writing. This concern motivates the investigation of methods that can reveal what evidence models actually use .

\subsection{Linguistic and Stylometric Feature-Based Detection}

Linguistic feature-based detection provides a complementary strategy by grounding predictions in interpretable properties of text. Early work, such as GLTR, demonstrated that machine-generated text tends to rely on high-probability tokens \cite{gehrmann-etal-2019-gltr}. Visualising token predictability improved human detection performance, suggesting measurable stylistic differences between human and generated writing.

Subsequent research incorporated stylometric and linguistic features into machine learning classifiers. For example, StyloAI evaluated stylometric features across multiple classifiers and achieved strong performance on educational datasets \cite{opara2024styloai}. However, such studies typically evaluate performance within a single dataset and do not examine cross-domain behaviour.

A comprehensive survey by Terčon et al. \cite{tervcon2025linguistic} synthesised lexical, syntactic, discourse, and stylistic patterns associated with machine-generated text. The authors emphasise three observations: no single feature reliably distinguishes authorship, feature importance varies across generators and datasets, and generalisation remains a major challenge. These findings suggest that linguistic cues may be context-dependent rather than universal indicators of machine authorship, but empirical validation across benchmarks remains limited.

\subsection{Robustness, Generalisation, and Ensemble Methods}

Robustness to domain and generator shift has emerged as one of the most persistent challenges in AI-generated text detection. Multiple studies report substantial performance degradation when detectors trained on one dataset are applied to different domains or newly released models.

To address this, ensemble-based approaches have been explored. Results from the COLING 2025 shared tasks indicate that combining multiple fine-tuned models improves performance across languages \cite{wang-etal-2025-genai}. Mobin et al. show that ensembles of RoBERTa, XLM-RoBERTa, and BERT variants outperform individual models \cite{mobin-islam-2025-luxveri}. Similarly, Pudasaini et al. \cite{pudasaini:2025}, and Kristanto et al. \cite{kristanto2025theoretically} demonstrate improved robustness using ensemble architectures.

Nevertheless, these methods primarily improve predictive performance without clarifying what the models detect. If improved robustness still relies on dataset-specific cues, higher accuracy alone does not resolve questions about detector validity, particularly in high-stakes applications.

\subsection{Explainable AI for Text Detection}

Explainable AI (XAI) has been proposed as a way to increase trust in detection systems by providing insight into model decisions. Educators and institutions require evidence supporting classification decisions, especially when outcomes may affect academic evaluation.

Recent work applies post-hoc explanation techniques such as SHAP and LIME to text detectors. Masih et al. visualised token-level contributions in transformer-based models \cite{masih2025classifying}, while Najjar et al. analysed lexical features influencing attribution decisions \cite{najjar2025leveragingexplainableaillm}. These studies demonstrate that explanations can reveal patterns associated with machine-generated text.

However, most work uses explainability primarily for transparency or model validation. Few studies use explanations as a scientific instrument to analyse what signal detectors rely on across datasets. Consequently, explainability has rarely been used to test whether models detect authorship itself or only corpus-specific stylistic differences.

\subsection{Summary and Research Gap}

Existing literature demonstrates three important observations: detectors can achieve high benchmark accuracy, performance often degrades under domain shift, and explainability can reveal model behaviour. Yet these findings have rarely been combined into a single investigation.

The central open question, therefore, remains: do current detectors identify stable linguistic signatures of machine authorship, or do they exploit dataset-specific cues? This work addresses that question by systematically evaluating linguistic-feature-based detectors across multiple benchmarks and using instance-level explanations to analyse the evidence behind their predictions.

\section{Methodology}

This section describes the experimental pipeline used to build and analyse an interpretable detector of AI-generated text. Our goal is not only to measure predictive performance, but to assess whether models rely on stable indicators of machine authorship or on dataset-specific cues. We first define the document-level linguistic feature space, then describe the benchmark datasets and exploratory analysis used to characterise domain differences. Finally, we detail the feature selection strategy, model training, evaluation protocols (in-domain, cross-domain, and cross-generator), ensemble construction, and SHAP-based interpretability.

\subsection{Linguistic Feature Representation}

Each document is represented as a vector of interpretable linguistic features, designed to capture systematic differences between human-written and AI-Generated text. Unlike embedding-based approaches, all features are explicitly defined and computed at the document level, enabling direct inspection of the linguistic properties that drive model predictions.

We extract 38 features drawn from established feature-based detection frameworks~\cite{opara2024styloai} and surveys of the linguistic characteristics of AI-Generated language~\cite{tervcon2025linguistic}. These features were selected to span a broad range of linguistic dimensions, ensuring that the representation is not biased towards any single aspect of writing style. Specifically, the feature space encompasses five complementary dimensions: (i) surface and length statistics, (ii) lexical diversity and distribution, (iii) syntactic and grammatical structure, (iv) readability and predictability proxies, and (v) discourse, style, sentiment, and expressivity signals. This multi-dimensional design is motivated by prior evidence that detection cues are distributed across multiple linguistic levels rather than concentrated in any single marker~\cite{tervcon2025linguistic}. Table~\ref{feature_definitions_full} presents the full feature set along with operational definitions and the corresponding calculation procedure for each feature.

To identify the most discriminative subset, Recursive Feature Elimination (RFE) was applied to the initial pool of 38 features, yielding an optimal set of 30 features for downstream classification. Features based on raw counts — such as character count, word count, and paragraph count — were normalised prior to model training using standard scaling, which transforms each feature to zero mean and unit variance. All features are computed consistently across datasets, ensuring that model decisions are grounded in transparent, reproducible linguistic signals rather than latent representations learned by large language models.

\begin{table*}[htbp]
\caption{Linguistic feature set with operational definitions and computational specifications. Sentence tokenisation uses NLTK Punkt \cite{bird2009natural}; POS tagging, lemmatisation, and dependency parsing use spaCy \texttt{en\_core\_web\_sm} \cite{honnibal2020spacy}; grammar checking uses LanguageTool v5.7 \cite{naber2003rule}; sentiment analysis uses TextBlob \cite{loria2018textblob}; TF-IDF vectorisation uses scikit-learn \cite{pedregosa2011scikit}.}
\label{feature_definitions_full}
\centering
\small
\setlength{\tabcolsep}{6pt}
\renewcommand{\arraystretch}{1.2}
\rowcolors{2}{gray!10}{white}
\begin{tabularx}{\textwidth}{@{}p{3.5cm}p{4.5cm}X@{}}
\toprule
\textbf{Feature} & \textbf{Description} & \textbf{How It Is Calculated} \\
\midrule

\multicolumn{3}{l}{\textbf{Surface and Length Features}} \\
Character Count & Total characters in the document. & Count all characters, including spaces. \\
Word Count & Total word tokens. & Split on whitespace and count tokens. \\
Sentence Count & Number of Sentences. & Sentence tokenisation count. \\
Paragraph Count & Number of paragraphs. & Count non-empty segments split by newlines. \\
Punctuation Count & Total punctuation symbols. & Count standard punctuation characters. \\

\midrule
\multicolumn{3}{l}{\textbf{Lexical Diversity and Distribution}} \\
Unique Word Count & Distinct word types. & Count unique lowercase tokens. \\
Vocabulary Size & Effective Vocabulary Size. & Count distinct lemmas. \\
Type--Token Ratio & Lexical diversity. & Unique words divided by total words. \\
Hapax Ratio & Words occurring exactly once. & Hapax count divided by total unique words. \\
Word Entropy & Lexical unpredictability. & Shannon entropy of lemma frequency distribution. \\
Repetition Rate & Repeated phrase usage. & Repeated bigrams divided by total bigrams. \\

\midrule
\multicolumn{3}{l}{\textbf{Readability and Predictability}} \\
Flesch Reading Ease & Ease of comprehension. & Standard Flesch formula via \texttt{textstat}. \\
Gzip Ratio & Textual Redundancy. & Compressed byte length divided by original length. \\
Predictability Score & Lexical Surprisal. & Mean negative log unigram probability from within-document frequencies. \\

\midrule
\multicolumn{3}{l}{\textbf{Syntactic and Grammatical Structure}} \\
Sentence Complexity & Syntactic complexity. & Average dependency depth per sentence. \\
Clause--Sentence Ratio & Clauses per sentence. & Clausal dependency count divided by sentence count. \\
Sentence Length Variation & Variability in sentence lengths. & Standard deviation of per-sentence word counts. \\
Sentence Length Difference & Range of Sentence Lengths. & Maximum minus minimum sentence word count. \\
POS Diversity & Part-of-speech diversity. & Shannon entropy of POS tag distribution. \\
POS Bigram Variety & POS Bigram Pattern Variety. & Count of distinct POS bigram types. \\
POS Trigram Variety & POS Trigram Pattern Variety. & Count of distinct POS trigram types. \\
POS 4-gram Variety & POS 4-gram pattern variety. & Count of distinct POS 4-gram types. \\
Grammatical Mistakes & Detected grammatical errors. & Total issues flagged by LanguageTool v5.7. \\

\midrule
\multicolumn{3}{l}{\textbf{Discourse, Style, Pragmatics, and Sentiment}} \\
Stopword Count & Stopwords present. & Count of tokens in the NLTK stopword list. \\
Discourse Marker Count & Discourse Marker Frequency. & Count from a predefined marker list. \\
Transition Variety Score & Connective diversity. & Count from a curated transition phrase list. \\
Paragraph Coherence & Logical flow across paragraphs. & Mean cosine similarity of adjacent paragraph TF-IDF vectors. \\
Pronoun Ratio & Pronoun Proportion. & Pronoun count divided by total words. \\
Personal Voice Score & First-person narrative style. & First-person pronoun count divided by total words. \\
Modal Frequency & Modal Verb Frequency. & Modal verb count divided by total words. \\
Negation Frequency & Negation term frequency. & Negation token count divided by total words. \\
Question--Statement Ratio & Interrogative vs.\ declarative ratio. & Number of questions divided by declarative sentences. \\
Hedge / Uncertainty Score & Hedging expressions. & Count from a curated hedge word list. \\
Sentiment Polarity & Positive vs Negative tone. & TextBlob polarity score ($-1$ to $+1$). \\
Sentiment Subjectivity & Subjective vs.\ objective language. & TextBlob subjectivity score ($0$ to $1$). \\
Emotion Variation & Affective shifts across text. & Mean absolute polarity difference between consecutive sentences. \\
Specificity Score & Concreteness of language. & Proportion of noun and numeral tokens. \\
Figurative Language Score & Figurative Expression Presence. & Count from a curated figurative marker list. \\

\bottomrule
\end{tabularx}
\end{table*}

\subsection{Datasets}

We use three datasets to evaluate effectiveness, robustness, and generalisability. Two benchmark corpora are used for training and in-domain evaluation, and a third independent dataset is used exclusively for cross-corpus testing.

For training and primary evaluation, we employ: (i) the PAN CLEF dataset from the Voight--Kampff Generative AI Detection task at PAN@CLEF 2025 \cite{bevendorff:2025}, and (ii) the COLING dataset released for the GenAI Content Detection task at COLING 2025 \cite{wang-etal-2025-genai}. To evaluate cross-corpus behaviour without contaminating training or feature selection, we test ensemble models on the Ghostbuster dataset \cite{verma2024ghostbusterdetectingtextghostwritten}.

These datasets differ in domain coverage, scale, and generation protocols. Such differences are central to our research question because they induce domain and generator shift, enabling us to test whether detection relies on transferable cues or on corpus-specific artefacts. Table~\ref{tab:dataset_comparison} summarises dataset size, domains, generators, and official metrics.

\begin{table}[htbp]
\caption{Comparison of the benchmark datasets used for the experiment with their data size and diversity details.}
\label{tab:dataset_comparison}
\centering
\small
\setlength{\tabcolsep}{6pt}
\renewcommand{\arraystretch}{1.15}
\rowcolors{2}{gray!10}{white}

\begin{tabularx}{\textwidth}{@{}>{\bfseries}p{2.9cm} >{\RaggedRight\arraybackslash}X >{\RaggedRight\arraybackslash}X >{\RaggedRight\arraybackslash}X@{}}
\toprule
\textbf{Attribute} & \textbf{PAN CLEF Dataset} & \textbf{COLING Dataset} & \textbf{Ghostbuster Dataset} \\
\midrule
Task Name
& Voight--Kampff Generative AI Detection 2025 (Subtask~1)
& GenAI Content Detection Task~1 (English \& Multilingual)
& Research benchmark (NAACL~2024) \\

Domains Considered
& Essays, news, fiction
& Peer reviews, student essays, scientific papers, news, and 8+ additional domains
& Student essays, creative writing, news \\

Training Data Size
& 23{,}707 samples (9{,}101 human / 14{,}606 AI)
& 610{,}767 English samples (228{,}922 human / 381{,}845 machine)
& Not used for training purposes \\

Dev / Validation Size
& 3{,}589 samples (1{,}277 human / 2{,}312 AI)
& 261{,}758 English samples (98{,}328 human / 163{,}430 machine)
& Same-prompt splits; 6{,}000 text(2000 human-written,2000 generated with GPT and 2000 generated with Claude); used only for evaluation \\

Generators
& GPT-4o and participant models with obfuscation
& GPT-4/4o, Mistral, Llama~3.1, Qwen-2, Claude, others
& GPT-3.5-turbo (training); Claude (generalisation) \\

Evaluation Metrics
& ROC--AUC, Brier, C@1, F1, F0.5u, mean score, FPR/FNR
& Macro-F1 (official), Accuracy, Micro-F1
& F1 score, Accuracy \\

Availability
& Zenodo via TIRA (restricted use)\footnotemark[1]
& Google Drive and Hugging Face\footnotemark[2]
& GitHub repository\footnotemark[3] \\
\bottomrule
\end{tabularx}

\footnotetext[1]{PAN CLEF dataset available via Zenodo: \url{https://zenodo.org/records/14962653}}
\footnotetext[2]{COLING dataset available via Hugging Face: \url{https://huggingface.co/datasets/Jinyan1/COLING_2025_MGT_en}}
\footnotetext[3]{Ghostbuster dataset available via GitHub: \url{https://github.com/vivek3141/ghostbuster-data}}
\end{table}

\subsubsection{Dataset Analysis}

Because linguistic features are sensitive to genre, domain, and writing conditions, detectors trained on a single benchmark may overfit dataset-specific stylistic patterns rather than learn generalisable indicators of machine authorship \cite{doughman2025exploring}. To quantify the extent of dataset shift before training, we conduct an exploratory analysis of feature distributions for human-written and AI-generated texts.

Table~\ref{tab:extended_feature_statistics} reports dataset-level means for optimal features obtained after RFE in both PAN CLEF and COLING datasets. The benchmarks show substantial differences in document length, lexical diversity, and stylistic properties, indicating a pronounced domain shift. PAN CLEF texts tend to be longer and more narrative, whereas COLING texts are shorter and more informational, with AI-generated samples often closer to human-written samples in aggregate statistics. These differences motivate our cross-domain and cross-generator evaluation settings.

\begin{table*}[htbp]
\centering
\caption{Average values of selected linguistic features for human-written and AI-generated texts across the PAN CLEF, COLING, and Ghostbuster benchmark datasets.}
\label{tab:extended_feature_statistics}
\small
\setlength{\tabcolsep}{5pt}
\renewcommand{\arraystretch}{1.15}
\rowcolors{2}{gray!8}{white}

\begin{tabularx}{\textwidth}{
@{}p{6.4cm}
S[table-format=4.3]
S[table-format=4.3]
S[table-format=4.3]
S[table-format=4.3]
S[table-format=4.3]
S[table-format=4.3]@{}
}
\toprule
\multirow{2}{*}{\textbf{Feature}} &
\multicolumn{2}{c}{\textbf{PAN CLEF}} &
\multicolumn{2}{c}{\textbf{COLING}} &
\multicolumn{2}{c}{\textbf{Ghostbuster}} \\
\cmidrule(lr){2-3} \cmidrule(lr){4-5} \cmidrule(lr){6-7}
& \textbf{Human} & \textbf{AI}
& \textbf{Human} & \textbf{AI}
& \textbf{Human} & \textbf{AI} \\
\midrule

Character Count              & 3941.826 & 3580.483 & 1527.844 & 1449.026 & 3535.748 & 3080.883 \\
Word Count                   & 695.456  & 574.973  & 259.964  & 237.821  & 586.937  & 475.910  \\
Paragraph Count              & 26.409   & 17.884   & 7.116    & 3.786    & 12.587   & 9.107    \\
Stopword Count               & 326.233  & 239.586  & 110.491  & 102.020  & 256.590  & 191.907  \\
Unique Word Count            & 381.030  & 332.886  & 109.093  & 104.482  & 215.613  & 192.170  \\
Sentiment Subjectivity       & 0.486    & 0.482    & 0.446    & 0.461    & 0.455    & 0.482    \\
Discourse Marker Count       & 0.664    & 0.509    & 0.383    & 0.551    & 2.125    & 1.377    \\
Sentence Complexity          & 21.496   & 19.893   & 20.060   & 20.972   & 17.149   & 17.101   \\
Punctuation Count            & 131.982  & 96.855   & 46.280   & 38.633   & 108.538  & 74.327   \\
Sentence Length Difference   & 57.165   & 35.093   & 30.478   & 25.437   & 31.037   & 25.055   \\
Type--Token Ratio            & 0.555    & 0.590    & 0.626    & 0.608    & 0.494    & 0.494    \\
Word Entropy                 & 5.053    & 5.025    & 5.902    & 5.953    & 6.751    & 6.760    \\
Flesch Reading Ease          & 64.261   & 43.255   & 78.218   & 72.069   & 82.070   & 69.808   \\
Gzip Ratio                   & 0.489    & 0.482    & 0.591    & 0.542    & 0.485    & 0.463    \\
Question--Statement Ratio    & 0.033    & 0.027    & 0.031    & 0.037    & 0.055    & 0.021    \\
Clause--Sentence Ratio       & 3.143    & 2.809    & 1.267    & 1.038    & 1.345    & 1.127    \\
Modal Frequency              & 0.015    & 0.009    & 0.012    & 0.014    & 0.012    & 0.011    \\
Pronoun Ratio                & 0.124    & 0.084    & 0.074    & 0.071    & 0.085    & 0.066    \\
POS Diversity                & 2.321    & 2.187    & 11.991   & 11.970   & 13.088   & 12.674   \\
Hapax Ratio                  & 0.681    & 0.722    & 0.477    & 0.461    & 0.353    & 0.355    \\
Figurative Language Score    & 1.983    & 2.020    & 0.707    & 0.633    & 1.446    & 1.155    \\
Grammatical Mistakes         & 14.075   & 6.410    & 5.344    & 4.478    & 5.765    & 4.373    \\
Paragraph Coherence Consistency & 0.198 & 0.247    & 0.052    & 0.077    & 0.222    & 0.210    \\
POS 2-gram Variety           & 108.721  & 86.007   & 57.956   & 55.932   & 86.647   & 76.409   \\
POS 3-gram Variety           & 319.918  & 235.901  & 128.710  & 120.660  & 237.765  & 201.605  \\
POS 4-gram Variety           & 518.771  & 387.350  & 188.288  & 172.916  & 393.843  & 328.312  \\
Repetition Rate              & 0.049    & 0.052    & 0.045    & 0.065    & 0.056    & 0.064    \\
Sentence Length Variation    & 14.695   & 8.580    & 8.904    & 7.898    & 7.568    & 6.109    \\
Specificity Score            & 0.207    & 0.264    & 0.257    & 0.263    & 0.232    & 0.261    \\
Transition Variety Score     & 0.535    & 0.718    & 0.294    & 0.511    & 1.473    & 1.459    \\

\bottomrule
\end{tabularx}

\vspace{0.3em}
\footnotesize
\emph{Note:} Values represent dataset-level averages computed over all documents.
All ratios are normalised per document.
\end{table*}

The divergence is consistent with dataset construction. PAN CLEF includes human-written texts drawn largely from fiction (e.g., Project Gutenberg) and authorship obfuscation corpora, where authors were instructed to disguise their writing style \cite{bevendorff:2025}. In contrast, COLING contains naturally occurring human-authored content from sources such as Wikipedia, Reddit, arXiv, and WikiHow \cite{wang-etal-2025-genai}, reflecting more conventional academic and instructional writing. This construction makes COLING a more challenging generalisation setting, as AI-generated text can more closely approximate typical informational prose.

\begin{figure}[htbp]
    \centering
    \includegraphics[width=0.8\linewidth]{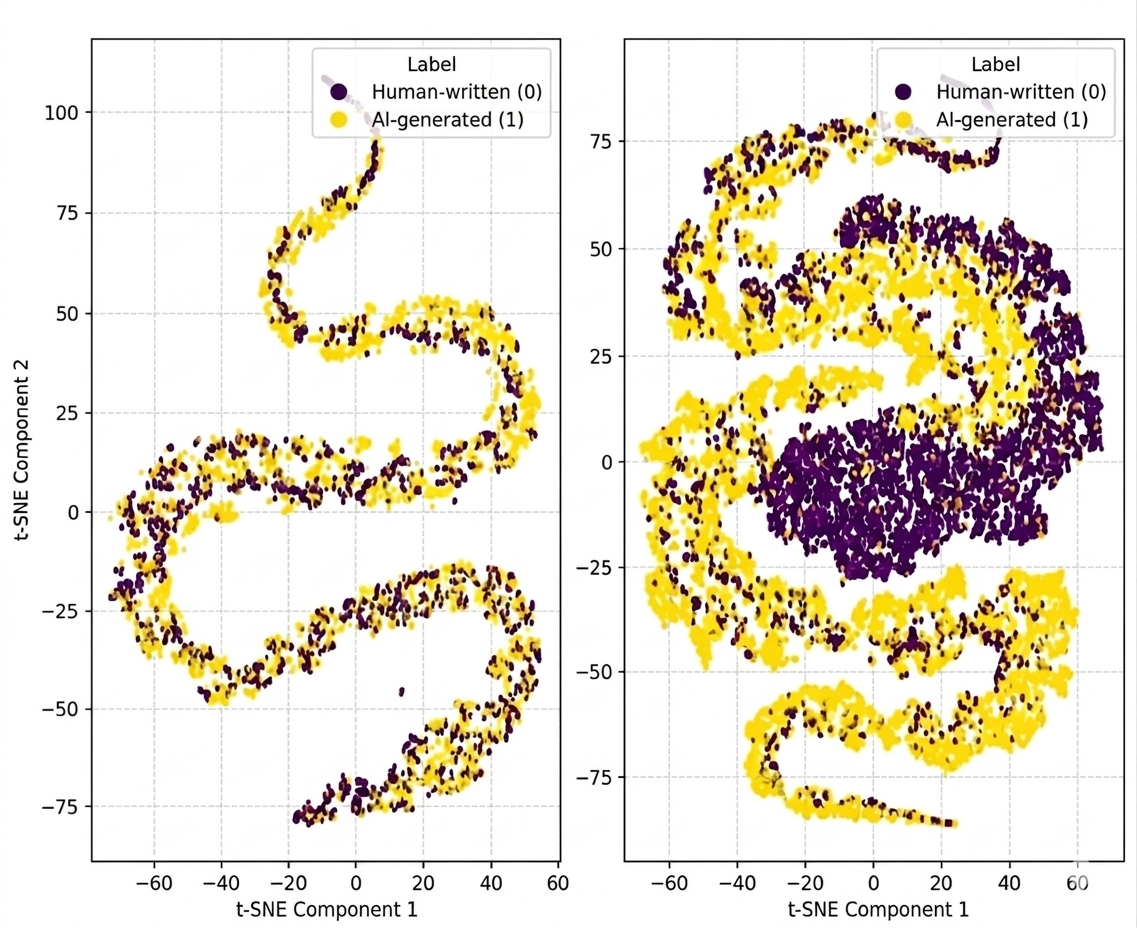}
    \caption{t-SNE projections of linguistic feature representations for the COLING dataset (left) and PAN CLEF dataset (right). Human-written samples are shown in purple and AI-generated samples in yellow.}
    \label{tsne_plot}
\end{figure}

To examine class separability in the feature space, we project the 38-dimensional representations into two dimensions using t-SNE. As shown in Figure~\ref{tsne_plot}, PAN CLEF exhibits partial class separation, while COLING shows substantial overlap. This visual evidence complements the quantitative results and suggests that feature-based discrimination may be less reliable under broader domain and generator diversity.

% (t-SNE figure unchanged)

\FloatBarrier

\subsection{Training and Evaluation}

\begin{figure}[htbp]
    \centering
    \includegraphics[width=0.9\linewidth]{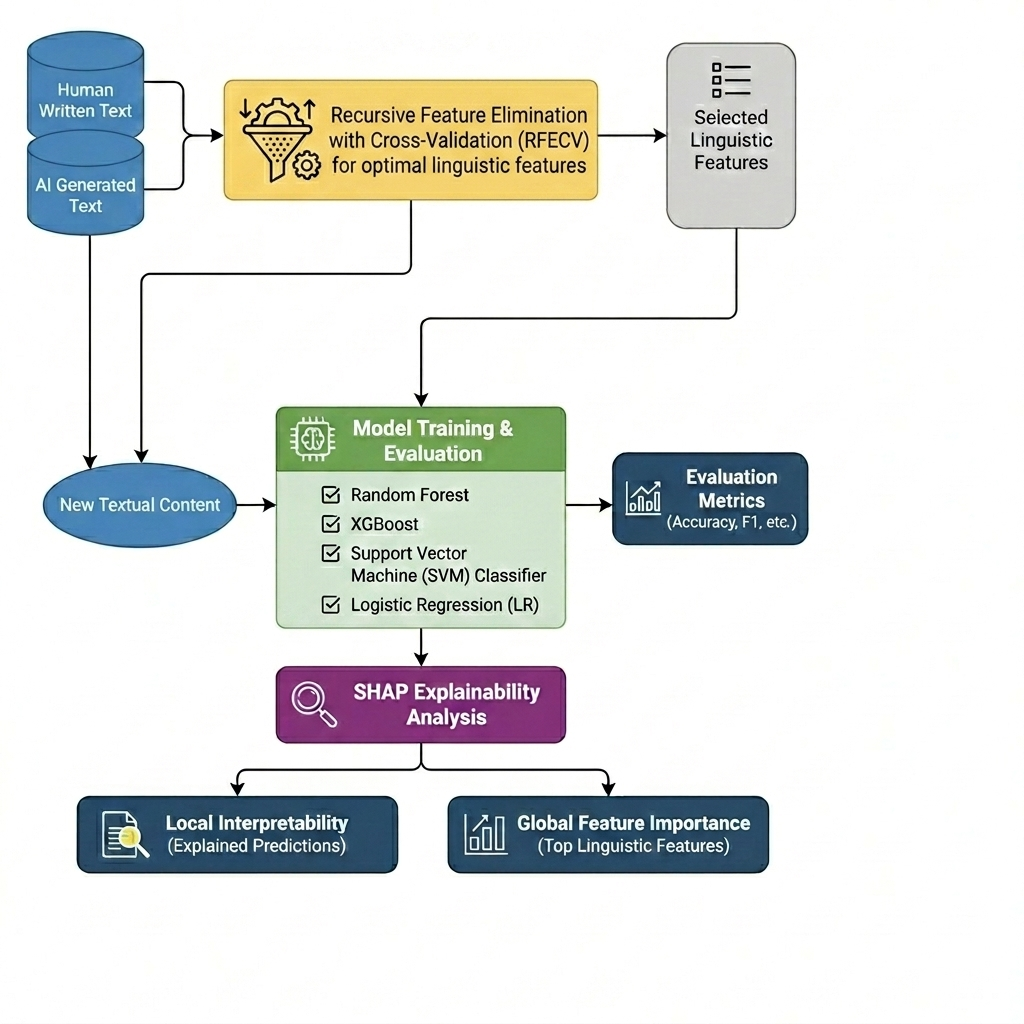}
    \caption{Overview of the experimental pipeline, illustrating dataset usage,
    linguistic feature extraction, model training and evaluation, ensemble
    construction, and post-hoc interpretability using SHAP.}
    \label{overall_diagram}
\end{figure}

Using the extracted features and benchmark datasets, we conduct experiments designed to answer three questions: (i) how accurate are feature-based models in-domain, (ii) how robust are they under domain and generator shift, and (iii) what linguistic evidence drives their predictions. Figure~\ref{overall_diagram} summarises the pipeline from feature extraction to evaluation and explainability.

% (Pipeline figure unchanged)

\paragraph{Feature selection.}
From the full pool of 38 features (Table~\ref{feature_definitions_full}, we apply Recursive Feature Elimination with Cross-Validation (RFECV) \cite{guyon2002gene} separately on PAN CLEF and COLING to obtain an optimal subset of 30 features per dataset. RFECV uses XGBoost as the base estimator and evaluates feature importance through 5-fold stratified cross-validation,  selecting subsets that maximise F1-score. Features are eliminated one 
at a time (step=1) to ensure granular selection. RFE iteratively removes weak or redundant predictors while preserving interpretability. Performing feature selection separately for each benchmark also supports our validity analysis: if different datasets yield different ``optimal'' feature subsets, this provides evidence that discriminative cues may be dataset-dependent rather than universal.

\paragraph{Base classifiers.}
We train four classifiers on the selected feature subsets: Logistic Regression, Support Vector Machine (SVM), Random Forest, and XGBoost. This selection enables comparison between linear models (Logistic Regression, SVM) and non-linear tree-based ensembles (Random Forest, XGBoost). Logistic Regression provides a strong, interpretable baseline for structured text features \cite{hassan2022analytics}. SVM is included for its effectiveness in high-dimensional feature spaces \cite{suthaharan2016support}. Random Forest and XGBoost are used because they capture non-linear interactions and typically perform well on tabular feature representations \cite{hua2024investigating}.

\paragraph{Evaluation settings.}
We evaluate models under three complementary conditions:
(i) \emph{in-domain} evaluation (train and test on the same benchmark),
(ii) \emph{cross-domain} evaluation (train on one benchmark and test on the other), and
(iii) \emph{cross-generator} evaluation (test on AI-generated samples produced by LLMs not present in training corpora).
Together, these settings distinguish benchmark performance from robustness to domain and generator shift.

\paragraph{Ensemble construction.}
To improve robustness, we construct an ensemble by averaging the predicted probabilities from the four base classifiers. Ensemble evaluation is performed on Ghostbuster \cite{verma2024ghostbusterdetectingtextghostwritten}, which is excluded from training and feature selection. For an input document $I$, each model outputs $P_k(I)$, the probability that $I$ is AI-generated. The final probability is computed using equal-weight averaging:

\begin{equation}
\label{eq:final_prediction}
P_{\text{Final}}(I) = \frac{P_{\text{XGB}}(I) + P_{\text{RF}}(I) + P_{\text{SVM}}(I) + P_{\text{LR}}(I)}{4}.
\end{equation}

A threshold of $0.5$ is used to assign class labels. We adopt equal weighting to avoid introducing additional tunable parameters and to preserve transparency in aggregation.

\paragraph{Post-hoc interpretability.}
To analyse model behaviour, we apply SHAP (SHapley Additive exPlanations) to all trained classifiers. SHAP provides both global and instance-level explanations by attributing each prediction to feature contributions. This enables inspection of which linguistic properties dominate model decisions, comparison of feature reliance across datasets, and diagnosis of failure modes under cross-domain and cross-generator settings, without altering the underlying models \cite{yan2025explainable}.

\FloatBarrier

\section{Experimental Results}

This section evaluates both the predictive performance and the validity of the proposed detection framework.
% Our objective is not only to determine how accurately models classify benchmark data, but to understand whether they detect machine authorship or instead rely on dataset-specific signals. 
Our primary objective is to investigate whether models trained on linguistic features genuinely detect machine authorship or merely exploit dataset-specific signals. To this end, we first establish that our classifiers achieve competitive in-domain performance, confirming that the models are sufficiently capable to make the subsequent cross-domain and cross-generator analysis meaningful.
We therefore analyse five complementary aspects: (i) in-domain benchmark performance, (ii) cross-dataset generalisation, (iii) robustness to unseen generators, (iv) ensemble behaviour on an independent corpus, and (v) explanations of model decisions using SHAP.

\subsection{In-domain Performance and Benchmark Comparison}

We first evaluate model performance under in-domain conditions across two benchmarks: the PAN CLEF 2025 Voight--Kampff Generative AI Detection shared task, and the COLING 2025 Task 1: Binary Multilingual Machine-Generated Text Detection shared task. For each benchmark, classifiers are trained on the respective training partition using the dataset-specific optimal feature subset identified via Recursive Feature Elimination with Cross-Validation (RFECV). Results are compared against official shared task baselines and top-ranked systems, as reported on the respective competition leaderboards (Table~\ref{tab:fscore_pan}).

Our feature-based machine learning models surpass all official baselines on both benchmarks. On the PAN CLEF 2025 benchmark, the best-performing model achieves an F1 score of 0.9733, placing second overall on the Voight--Kampff Generative AI Detection leaderboard. On the COLING 2025 benchmark, the trained classifier achieves an F1 score of 80.25, ranking seventh in the Binary Multilingual Machine-Generated Text Detection task. Notably, while the top-ranked system~\cite{macko2025mdok} leverages fine-tuned large language models and supplementary training data, our approach attains competitive performance using only interpretable linguistic features and conventional machine learning algorithms. These results confirm that our classifiers are sufficiently capable to serve as a meaningful basis for the cross-domain and cross-generator generalisation analysis that follows.

% We first evaluate performance under in-domain conditions using both the PAN CLEF 2025 benchmark and the COLING 2025 Task 1: Binary Multilingual Machine-Generated Text Detection benchmark. Classifiers are trained on the respective dataset training set using the dataset-specific optimal feature subset obtained via Recursive Feature Elimination with Cross-Validation (RFECV). Performance is compared against official baselines and top-ranked systems from the Voight--Kampff Generative AI Detection 2025 leaderboard (Table~\ref{tab:fscore_pan}).

% Our feature-based ML models outperform all official baselines and rank second overall, achieving an F1 score of 0.9733 in the PAN CLEF 2025 benchmark whereas the COLING dataset trained model achieved an F1 score of 80.25 ranking us in the 7th of the Task 1: Binary Multilingual Machine-Generated Text Detection competition. While the top-ranked system \cite{macko2025mdok} uses fine-tuned large language models and additional training data, our approach reaches comparable performance using interpretable linguistic features and conventional machine learning.

However, in-domain accuracy alone does not demonstrate reliable authorship detection. Because training and testing samples originate from the same dataset, models may learn corpus-specific stylistic patterns rather than general properties of machine-generated language. The following experiments, therefore, test whether this performance transfers beyond the benchmark setting.

\begin{table}[htbp]
\centering
\caption{Comparison of F-score performance on the Voight--Kampff Generative AI Detection 2025 leaderboard (F-score scaled to 0--100).}
\label{tab:fscore_pan}
\begin{tabular}{lc}
\hline
\textbf{Algorithm} & \textbf{F1-score} \\
\hline
mdok of KInIT \cite{macko2025mdok} & 98.90 \\

\textbf{Optimised Feature Set + SVC (Ours)} & \textbf{97.34} \\
\textbf{Optimised Feature Set + XGBoost (Ours)} & \textbf{96.94} \\
\textbf{Optimised Feature Set + Random Forest (Ours)} & \textbf{95.82} \\
\textbf{Optimised Feature Set + Logistic Regression (Ours)} & \textbf{95.48} \\
Baseline TF-IDF SVM \cite{bevendorff:2025} & 90.40 \\
Baseline Binoculars LLaMA 3.1 \cite{bevendorff:2025} & 86.60 \\
\hline
\end{tabular}
\end{table}

While the F1-score summarises overall predictive performance, it does not reveal the type of errors made by the detector. In high-stakes educational settings, the most critical failure is a false positive (human-written text incorrectly classified as AI-generated), as this may lead to unjust academic accusations. 
The full report of the models trained is shown in the Table \ref{tab:model_performance_comparison}.

\begin{table}[htbp]
\centering
\caption{Comparison of machine learning model performance on the AI-generated text detection dataset using cross-validation. Reported metrics include Accuracy, Precision, Recall, F1-score, False Positive Rate (FPR), and False Negative Rate (FNR). The best-performing model across most evaluation metrics is highlighted in bold. All values are scaled from 0 to 100.}
\label{tab:model_performance_comparison}
\begin{tabular}{lcccccc}
\hline
\textbf{Model} & \textbf{Accuracy} & \textbf{Precision} & \textbf{Recall} & \textbf{F1 Score} & \textbf{FPR} & \textbf{FNR} \\
\hline
XGBoost & 96.04 & 96.45 & 97.45 & 96.94 & 6.50 & 2.55 \\
Random Forest & 94.51 & 94.12 & 97.58 & 95.82 & 11.04 & 2.42 \\
Logistic Regression & 94.09 & 94.19 & 96.80 & 95.48 & 10.08 & 3.20 \\
\textbf{SVM} & \textbf{96.55} & \textbf{96.71} & \textbf{97.97} & \textbf{97.34} & \textbf{6.03} & \textbf{2.03} \\
\hline
\end{tabular}
\end{table}

\begin{figure}[htbp]
\centering
\includegraphics[width=0.9\textwidth]{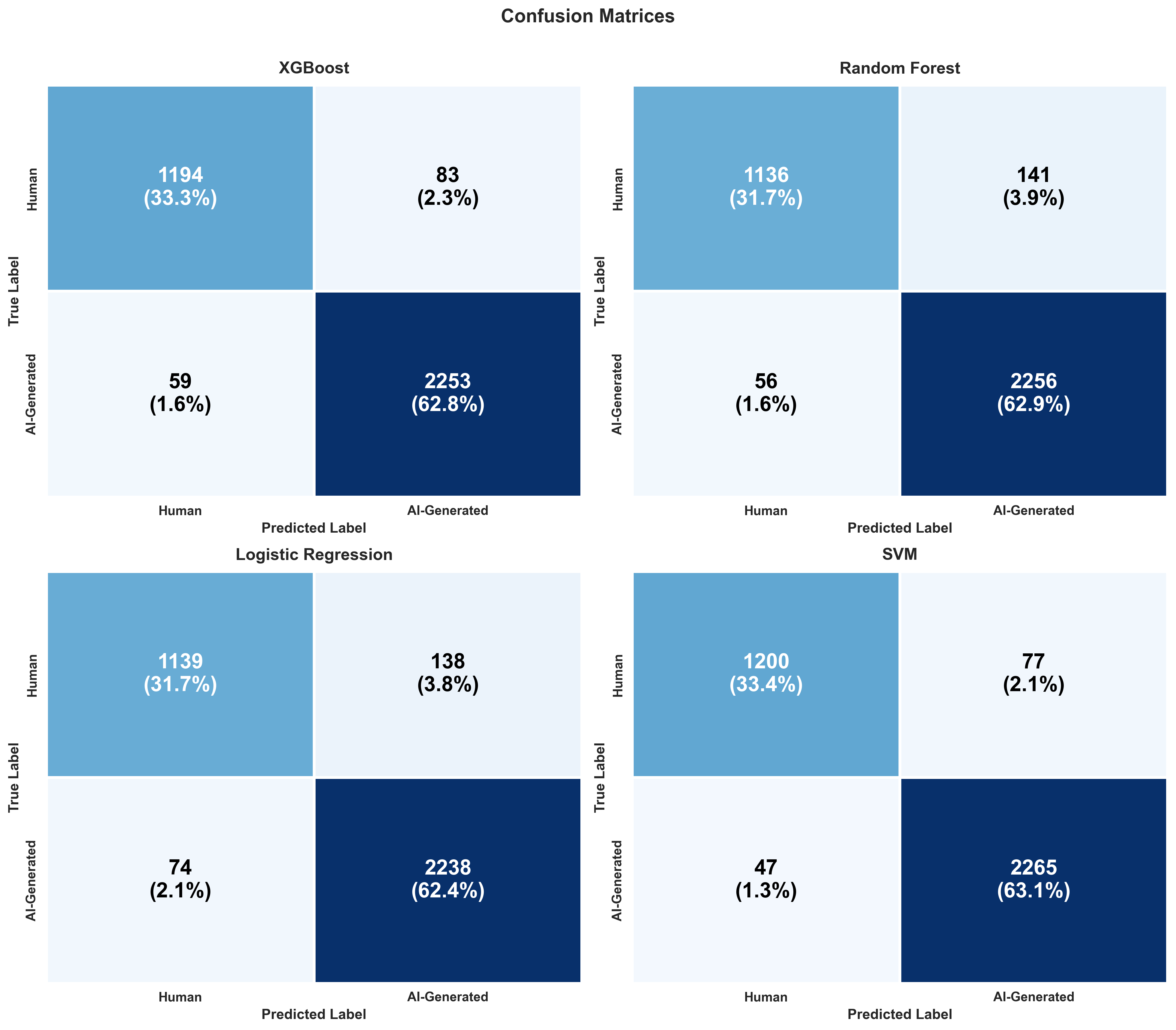}
\caption{Confusion matrices for the four classifiers evaluated on the PAN CLEF in-domain test set. Rows indicate true labels and columns predicted labels. False positives (Human $\rightarrow$ AI) represent potential false accusations, while false negatives (AI $\rightarrow$ Human) represent missed detections. Despite similar F1 scores, models differ in their error distribution, illustrating why benchmark accuracy alone does not fully capture deployment risk.}
\label{fig:pan_confusion_matrices}
\end{figure}

Additionally, we analyse the confusion matrices of all classifiers to expose the balance between false positives (human-written text misclassified as AI) and false negatives (AI-Generated text misclassified as human-written). 
Figure~\ref{fig:pan_confusion_matrices} visualises prediction behaviour for the PAN CLEF test set.

Across models, false negatives remain relatively low, indicating that AI-generated text is usually detected. However, false positives occur non-negligibly, meaning a proportion of genuine student work would be flagged as AI-written even under in-domain conditions. This observation is important: a detector can achieve near state-of-the-art F1 while still posing a practical risk if deployed without human review. Consequently, accuracy alone is insufficient to justify operational use, and error asymmetry must be considered alongside benchmark scores.

\subsection{Validation of models}
Model robustness was assessed through stratified 10-fold cross-validation, and pairwise statistical significance was established through both paired t-tests at $\alpha = 0.05$.
Stratified 10-fold cross-validation confirmed the stability and generalisability of the observed rankings. The mean and standard deviation of key metrics across folds are reported in the Table \ref{tab:model_performance_comparison_cv}.

\begin{table}[htbp]
\centering
\caption{Comparison of machine learning model performance using cross-validation. Reported metrics include Accuracy, Precision, Recall, F1 Score, and ROC-AUC. The best-performing model across most evaluation metrics is highlighted in bold. Values are reported as mean $\pm$ standard deviation.}
\label{tab:model_performance_comparison_cv}
\begin{tabular}{lccccc}
\hline
\textbf{Model} & \textbf{Accuracy} & \textbf{Precision} & \textbf{Recall} & \textbf{F1 Score} & \textbf{ROC-AUC} \\
\hline
\textbf{SVM} & \textbf{0.9697 $\pm$ 0.0045} & \textbf{0.9731 $\pm$ 0.0051} & \textbf{0.9778 $\pm$ 0.0036} & \textbf{0.9754 $\pm$ 0.0036} & \textbf{0.9942 $\pm$ 0.0015} \\
XGBoost & 0.9678 $\pm$ 0.0033 & 0.9717 $\pm$ 0.0028 & 0.9762 $\pm$ 0.0034 & 0.9739 $\pm$ 0.0027 & 0.9943 $\pm$ 0.0009 \\
Random Forest & 0.9505 $\pm$ 0.0044 & 0.9500 $\pm$ 0.0037 & 0.9707 $\pm$ 0.0044 & 0.9602 $\pm$ 0.0035 & 0.9878 $\pm$ 0.0019 \\
Logistic Regression & 0.9431 $\pm$ 0.0043 & 0.9490 $\pm$ 0.0048 & 0.9591 $\pm$ 0.0050 & 0.9540 $\pm$ 0.0034 & 0.9826 $\pm$ 0.0015 \\
\hline
\end{tabular}
\end{table}

The results in Table~\ref{tab:model_performance_comparison_cv} show that the standard deviations are consistently small ($\sigma \leq 0.005$ across all metrics), indicating low variance and robust generalisation across data splits.

For statistical validation, to determine the appropriate statistical framework for comparing classifier performance, we applied the Shapiro-Wilk test \cite{shaphiro1965analysis} to the per-fold scores of all four classifiers across five evaluation metrics (Accuracy, Precision, Recall, F1, and ROC-AUC). All 20 tests (4 models × 5 metrics) yielded p-values above the 0.05 significance level (W-statistics ranging from 0.9134 to 0.9882), confirming that all metrics follow a normal distribution. Since the Shapiro-Wilk test confirmed normality across all model-metric combinations, we employed a paired t-test to evaluate the statistical significance of performance differences between classifiers, which are reported in the Table \ref{tab:statistical_significance_models}. 

\begin{table}[htbp]
\centering
\caption{Statistical significance testing of F1-score differences between machine learning models using paired t-test. $\Delta$F1 indicates the difference in mean F1-score between the compared models. Statistical significance is determined at $\alpha = 0.05$.}
\label{tab:statistical_significance_models}
\begin{tabular}{lccc}
\hline
\textbf{Comparison} & \textbf{$\Delta$F1} & \textbf{t-test p-value} & \textbf{Significant?} \\
\hline
SVM vs. XGBoost & +0.0015 & 0.0984 & No \\
SVM vs. Random Forest & +0.0152 & $< 0.0001$ & Yes \\
SVM vs. Logistic Regression & +0.0214 & $< 0.0001$ & Yes \\
XGBoost vs. Random Forest & +0.0137 & $< 0.0001$ & Yes \\
XGBoost vs. Logistic Regression & +0.0199 & $< 0.0001$ & Yes \\
Random Forest vs. Logistic Regression & +0.0062 & $< 0.0001$ & Yes \\
\hline
\end{tabular}
\end{table}

Results from Table~\ref{tab:statistical_significance_models} reveal a clear two-tier structure. SVM and XGBoost form a statistically indistinguishable top tier ($p > 0.05$ ), while both significantly outperform Random Forest and Logistic Regression. All other pairwise differences are statistically significant at $\alpha = 0.05$, confirmed by a paired t-test. The failure to reject the null hypothesis for SVM vs.\ XGBoost ($p = 0.098$ for the t-test) is an important result. Rather than declaring a single best model, the analysis supports the conclusion that both SVM and XGBoost achieve statistically equivalent performance on this task.

% \subsection{Cross-dataset Generalisation}

\subsection{Cross-Domain Generalisation}

Table~\ref{tab:cross_dataset_results} reports cross-dataset generalisation results for all classifiers trained on one benchmark and evaluated on the other. A pronounced asymmetry is observed in transfer behaviour depending on the direction of generalisation.

Models trained on PAN CLEF exhibit a substantial decline when applied to COLING. For instance, XGBoost drops from an in-domain F1 of 96.94 to 67.23 on COLING, and similar degradation is observed across all classifiers. This suggests that models trained on PAN CLEF capture distributional properties specific to that dataset, which do not transfer reliably to the more diverse COLING setting.

The reverse direction tells a notably different story. COLING-trained models applied to PAN CLEF achieve F1 scores ranging from 78.05 to 82.84 — scores that are comparable to, and in several cases marginally above, their own in-domain COLING performance. This pattern has two plausible interpretations that are not mutually exclusive. First, COLING may represent a more linguistically diverse training distribution, producing models that generalise more robustly. Second, and importantly, the lower absolute performance on COLING — approximately 80\% in-domain compared to over 95\% in-domain on PAN CLEF — may reflect that COLING is an intrinsically more challenging task, encompassing a broader range of generators, languages, and domains. Under this interpretation, the performance gap between the two benchmarks is at least partially a property of task difficulty rather than model failure.

Taken together, these results indicate that strong in-domain benchmark accuracy is not a reliable indicator of cross-domain robustness, and that the composition and diversity of the training distribution play a critical role in determining generalisation behaviour.

% To evaluate robustness under domain shift, we train models on one benchmark dataset and test them on the other (Table~\ref{tab:cross_dataset_results}). This setting approximates real deployment scenarios, where detectors encounter unseen domains and writing styles.

% A substantial performance drop occurs when models trained on PAN CLEF are applied to COLING. For example, XGBoost decreases from 96.94 F1 in-domain to 67.23 F1 cross-domain. Similar degradation is observed across all classifiers. In contrast, models trained on the more diverse COLING dataset transfer better to PAN CLEF, although still below in-domain performance.

% These results indicate that high benchmark accuracy depends strongly on dataset characteristics. Tree-based models achieve the highest in-domain performance but exhibit the largest degradation, suggesting that they capture dataset-specific patterns. Linear models show lower peak accuracy but more stable cross-domain behaviour. 
% The findings reveal a trade-off between optimisation for benchmark performance and generalisation to new data.

\begin{table}[htbp]
\centering
\caption{Cross-dataset generalisation results on PAN CLEF and COLING test sets (F1 scaled to 1--100).}
\label{tab:cross_dataset_results}
\begin{tabular}{l l c c}
\hline
\textbf{Train Set} & \textbf{Algorithm} & \textbf{PAN CLEF} & \textbf{COLING} \\
\hline
\multirow{4}{*}{PAN CLEF}
& SVC & \textbf{97.34} & 76.59 \\
& XGBoost & 96.94 & 67.23 \\
& Random Forest & 95.82 & 74.86 \\
& Logistic Regression & 95.48 & 76.53 \\
\hline
\multirow{4}{*}{COLING}

& SVC & 78.05 & 76.79 \\
& XGBoost & 78.56 & 78.53 \\
& Random Forest & 79.95 & \textbf{80.25} \\
& Logistic Regression & 82.84 & 75.50 \\

\hline
\end{tabular}
\end{table}

\subsection{Cross-Generator Robustness}
To evaluate robustness to unseen language models, we test our trained classifiers on AI-generated text produced by generators entirely absent from all training corpora. The evaluation set comprises 1,049 samples from DeepSeek V3.2, 761 samples from GPT-5.2, and 228 samples from Gemini 3 Pro, all generated via the respective vendors' Application Programming Interface (API) services. These models were deliberately selected on the basis that they were released after the curation of both training datasets, thereby eliminating any risk of overlap between training and test distributions. Since this evaluation set contains only AI-generated text, performance is measured exclusively by the False Negative Rate (FNR) — the proportion of AI-generated text samples incorrectly classified as human-written text as reported in Table~\ref{tab:llm_generalization}.

% The results reveal considerable variability in cross-generator performance, indicating that neither model generalises reliably to unseen language models. The Best-PAN model achieves a low FNR on DeepSeek and Gemini, yet fails substantially on GPT-5.2. Conversely, the Best-COL model performs well on GPT-5.2 and DeepSeek but exhibits a markedly elevated FNR on Gemini. This inconsistency across generators, with no single model demonstrating consistently robust detection, suggests that both classifiers have learned patterns that are specific to the generators represented in their respective training distributions rather than features that characterise AI-generated text more broadly. Furthermore, the observed failures appear to be related to the degree to which a given generator's output approximates human writing: when a model produces text whose linguistic properties fall closer to the human distribution, feature-based detection becomes increasingly unreliable. These findings underscore a fundamental limitation of the feature-based detection paradigm and reinforce the central argument of this paper — that strong benchmark performance does not imply genuine generalisation to the diverse and evolving landscape of real-world AI-generated text.

The results from Table~\ref{tab:llm_generalization} demonstrate substantial variability in cross-generator performance, highlighting poor generalisation across unseen LLMs. The Best-PAN model achieves low FNR on DeepSeek and Gemini but fails on GPT-5.2, while the Best-COL model performs well on GPT-5.2 and DeepSeek but shows a high FNR on Gemini.

This inconsistency indicates that both models rely on dataset-specific patterns rather than learning generalisable features of AI-generated text.  Instead, they appear sensitive to stylistic properties specific to particular generators. When a generator produces text closer to human distributions, feature-based detection becomes unreliable. These findings underscore a fundamental limitation of the feature-based detection paradigm and reinforce the central argument of this paper — that strong benchmark performance does not imply genuine generalisation to the diverse and evolving landscape of real-world AI-generated text.

\begin{table}[htbp]
\centering
% \caption{Cross-generator performance on unseen LLMs (False Negative Rate (\%)).}
\caption{Cross-generator generalisation performance on unseen LLMs, reported as False Negative Rate (FNR, \%). \textbf{Best-PAN} refers to the SVM model achieving the highest F1-score when trained and evaluated on the PAN CLEF dataset, while \textbf{Best-COL} denotes the Random Forest model achieving the highest F1-score when trained and evaluated on the COLING dataset.}
\label{tab:llm_generalization}
\begin{tabular}{l c c}
\hline
\textbf{LLM} & \textbf{Best-PAN} & \textbf{Best-COL} \\
\hline
GPT-5.2 & 23.25 & 3.42 \\
DeepSeek V3.2 & 0.76 & 3.51 \\
Gemini 3 Pro & 0.43 & 18.40 \\
\hline
\end{tabular}
\end{table}

\subsection{Ensemble Performance on Unseen Data}
% \subsection{Ensemble Generalisation on Ghostbuster}

To mitigate the weaknesses of individual classifiers, we evaluate ensemble models on the Ghostbuster dataset, which is excluded from both training and feature selection, providing a fully held-out test of generalisation (Table~\ref{tab:ghostbuster_results}). The ensemble combines SVC, Random Forest, XGBoost, and Logistic Regression via majority voting. Performance is reported separately for three domain subsets: Wikipedia articles (WP), Reuters news articles, and student essays.

Ensembling produces substantial and consistent improvements over the best individual classifiers across both training conditions. For PAN-trained models, the average F1 score increases from 64.0 to 94.61, and for COLING-trained models, from 78.59 to 87.13. These gains indicate that aggregating diverse decision boundaries yields considerably greater robustness to domain shift than any single classifier achieves in isolation.

Examining domain-level results more closely reveals an informative pattern. The most pronounced improvements are observed on Reuters and Essay subsets, where ensemble F1 scores reach as high as 96.88 and 94.09, respectively, suggesting that AI-generated text in these domains retains detectable linguistic regularities that complementary classifiers collectively capture. The Wikipedia subset presents a more challenging case: Ens-PAN achieves 92.87, and Ens-COL achieves 77.77, both representing a marked improvement over the corresponding individual classifiers (60.15 and 59.36), yet remaining below the performance observed on other domains. This relative difficulty likely reflects the encyclopaedic and highly structured nature of Wikipedia text, which may more closely approximate the factual, neutral register that LLMs are inclined to produce, thereby reducing the discriminative signal available to feature-based models.

Collectively, these results demonstrate that ensembling substantially reduces cross-domain performance variation, even if it does not fully equalise performance across all domains. Crucially, even in the most challenging domain, the ensemble considerably outperforms its individual components, reinforcing the practical value of classifier combination as a strategy for improving robustness under distribution shift.

% To mitigate individual model weaknesses, we evaluate ensemble classifiers on the Ghostbuster dataset, which is excluded from both training and feature selection (Table~\ref{tab:ghostbuster_results}) We evaluate the ensemble model performance on different domains of the ghostbuster dataset (Wikipedia text (WP), News Articles from Reuters (Reuters), and Essays written by students (Essay)). 

% Ensembling substantially improves performance, increasing average F1 from 64.0 to 94.61 for PAN-trained models and from 78.59 to 87.13 for COLING-trained models. This demonstrates that combining diverse decision boundaries increases robustness to domain shift. However, ensemble improvement does not eliminate cross-dataset variation, indicating that robustness gains arise from complementary errors rather than the detection of universal authorship cues.

\begin{table}[htbp]
\centering
% \caption{Ensemble performance on the Ghostbuster dataset (F1 score).}
\caption{Ensemble performance on the Ghostbuster dataset (F1-score). \textbf{Ens-PAN} and \textbf{Ens-COL} denote ensemble models combining SVM, Random Forest, XGBoost, and Logistic Regression, trained on the PAN CLEF and COLING datasets, respectively.}
\label{tab:ghostbuster_results}
\begin{tabular}{l c c c c}
\hline
\textbf{Subset} & Best-PAN & Best-COL & Ens-PAN & Ens-COL \\
\hline
WP & 60.15 & 59.36 & 92.87 & 77.77 \\
Reuters & 55.80 & 85.59 & 96.88 & 90.98 \\
Essay & 76.07 & 90.83 & 94.09 & 92.64 \\
\hline
Average & 64.0 & 78.59 & 94.61 & 87.13 \\
\hline
\end{tabular}
\end{table}

\subsection{Explainable AI Results}
The algorithms have a very high in-domain performance but are not robust to cross-dataset generalisation and cross-generator generalisation. To investigate this failure in generalisation and understand what the models actually learn, we analyse predictions using SHAP explanations. We examine both global feature importance and individual predictions.

\begin{figure}[htbp]
    \centering
    \begin{subfigure}{0.48\textwidth}
        \centering
        \includegraphics[width=\linewidth]{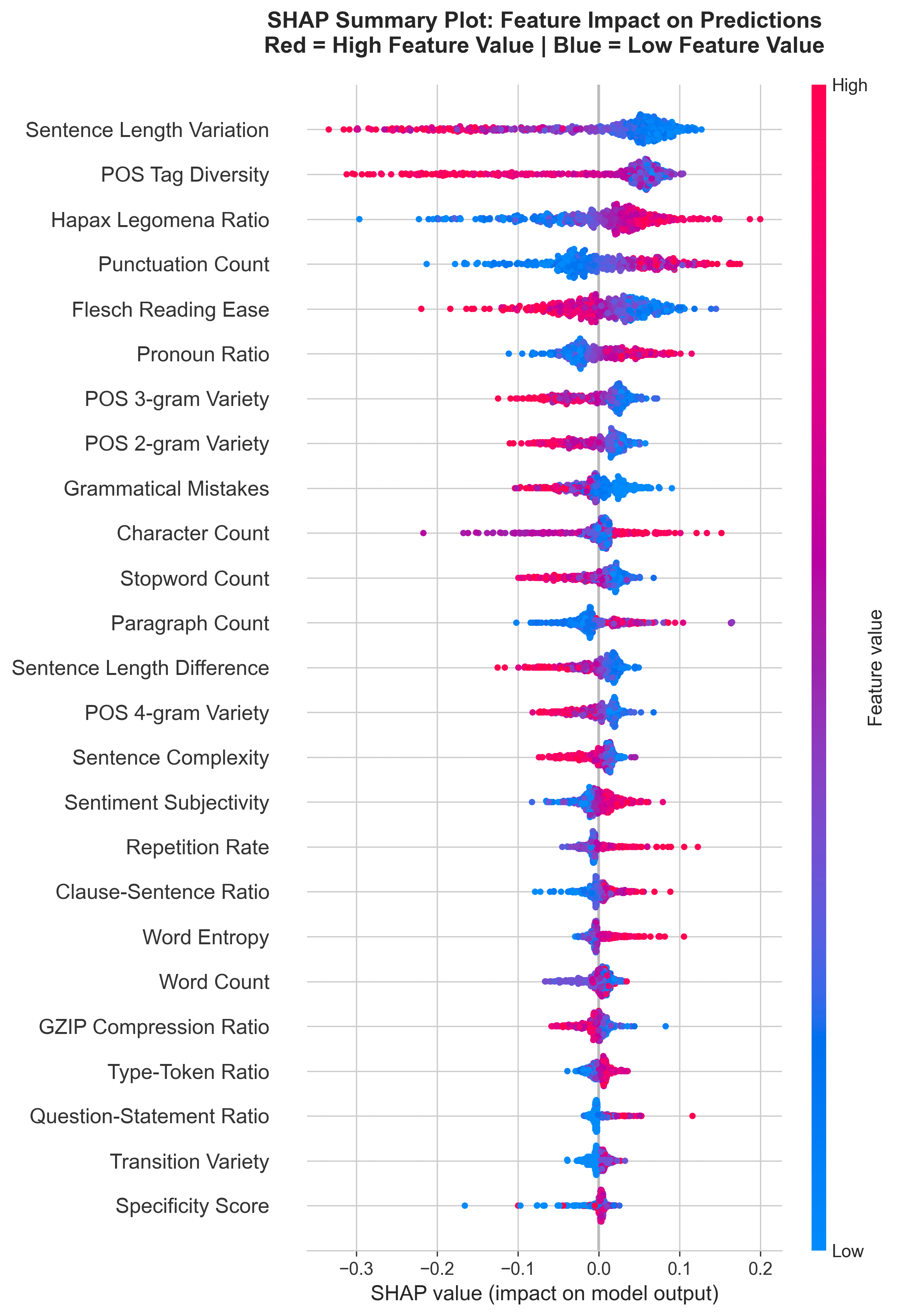}
        \caption{PAN CLEF dataset}
        \label{fig:shap_pan}
    \end{subfigure}\hfill
    \begin{subfigure}{0.48\textwidth}
        \centering
        \includegraphics[width=\linewidth]{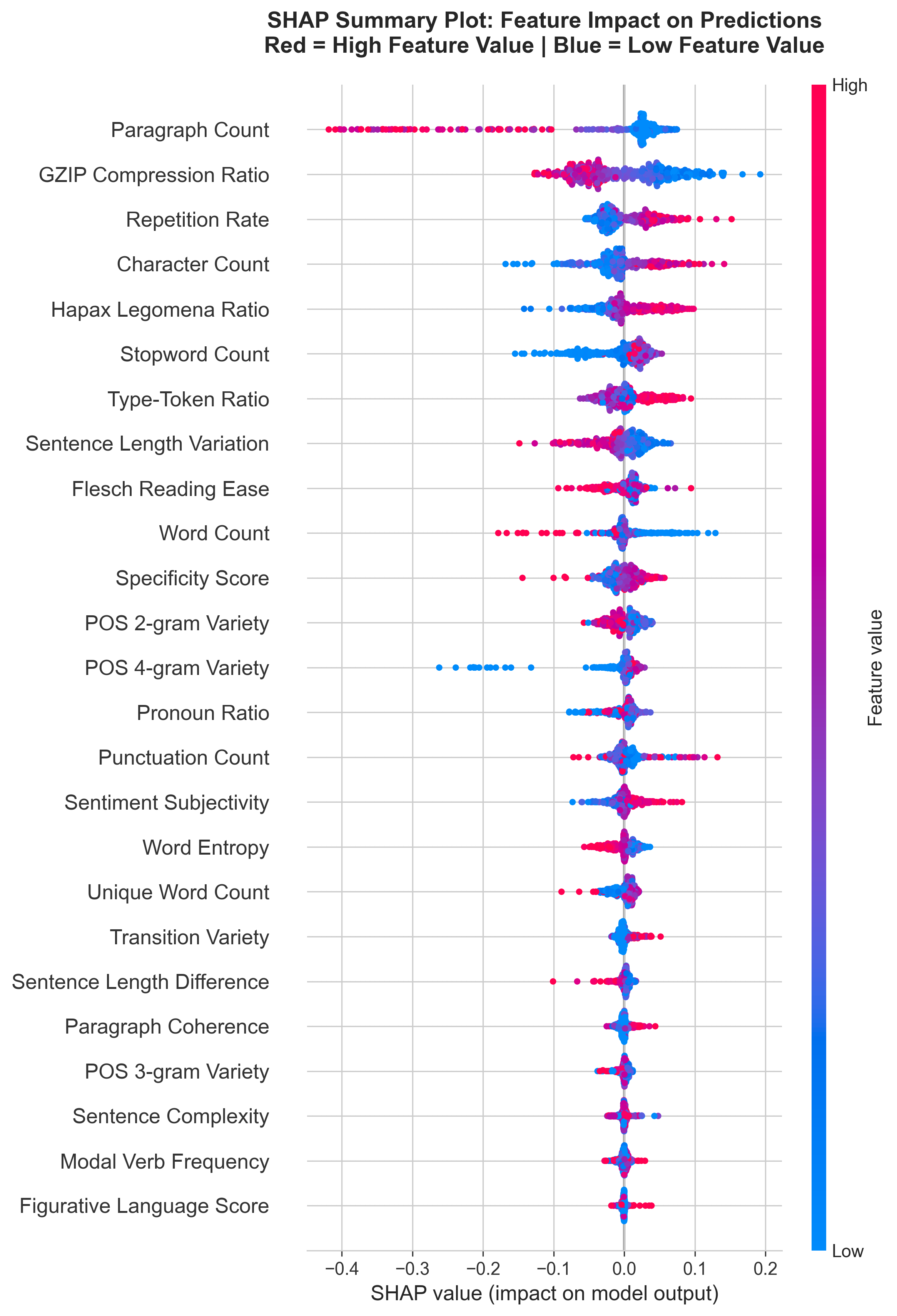}
        \caption{COLING dataset}
        \label{fig:shap_coling}
    \end{subfigure}
    \caption{SHAP summary plots showing feature importance distributions for different datasets.}
    \label{fig:shap_summary}
\end{figure}

Global explanations reveal striking differences between datasets. In the PAN CLEF model, Part-of-Speech diversity dominates feature importance (Figures~\ref{fig:feature_pan} and~\ref{fig:shap_pan}). In contrast, the COLING model relies primarily on structural and information-theoretic features such as Paragraph Count and GZIP compression ratio (Figures~\ref{fig:feature_coling} and~\ref{fig:shap_coling}).

% \begin{figure}[htbp]
%     \centering
%     \begin{subfigure}{0.5\textwidth}
%         \centering
%         \includegraphics[width=\linewidth]{Figures/PAN_feature_importance.png}
%         \caption{PAN CLEF dataset}
%         \label{fig:feature_pan}
%     \end{subfigure}\hfill
%     \begin{subfigure}{0.5\textwidth}
%         \centering
%         \includegraphics[width=\linewidth]{Figures/COLING_feature_importance.png}
%         \caption{COLING dataset}
%         \label{fig:feature_coling}
%     \end{subfigure}
%     \caption{Feature importance obtained while training an XGBoost model on different datasets.}
%     \label{fig:feature_importance}
% \end{figure}

% \begin{figure}[htbp]
%     \centering
%     \begin{subfigure}{0.9\textwidth}
%         \centering
%         \includegraphics[width=\linewidth]{Figures/PAN_feature_importance.png}
%         \caption{PAN CLEF dataset}
%     \end{subfigure}

%     \vspace{0.5cm}

%     \begin{subfigure}{0.9\textwidth}
%         \centering
%         \includegraphics[width=\linewidth]{Figures/COLING_feature_importance.png}
%         \caption{COLING dataset}
%     \end{subfigure}

%     \caption{Feature importance obtained while training an XGBoost model on different datasets.}
% \end{figure}

\begin{figure}
    \centering
    \includegraphics[width=0.75\linewidth]{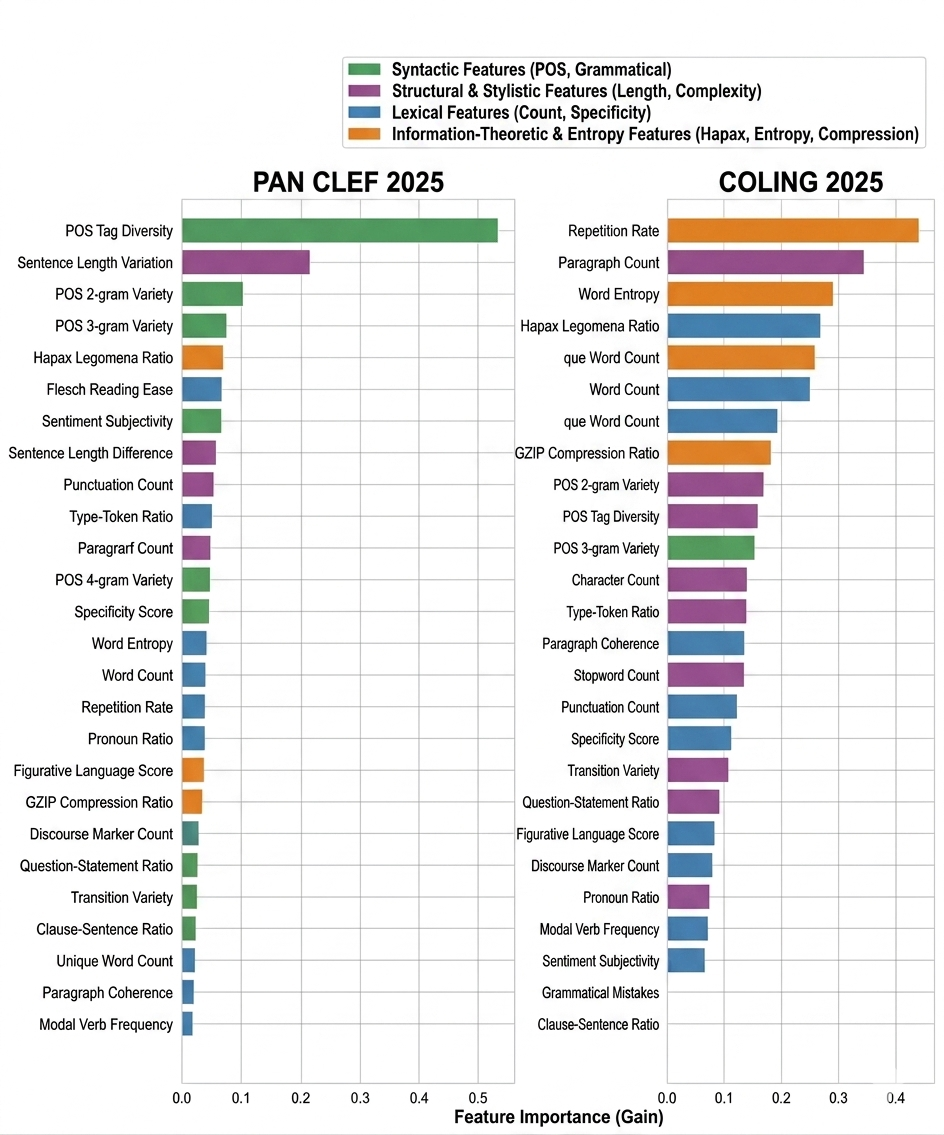}
    \caption{Feature importance obtained while training an XGBoost model on different datasets.}
    \label{fig:feature_importance}
\end{figure}

The minimal overlap in important features demonstrates that detection cues are dataset-dependent rather than universal. If models were detecting inherent properties of machine-generated language, feature importance would remain consistent across datasets. Instead, explanations show that different corpora induce different decision criteria.

\begin{figure}[htbp]
    \centering
    \begin{subfigure}{0.5\textwidth}
        \centering
        \includegraphics[width=\linewidth]{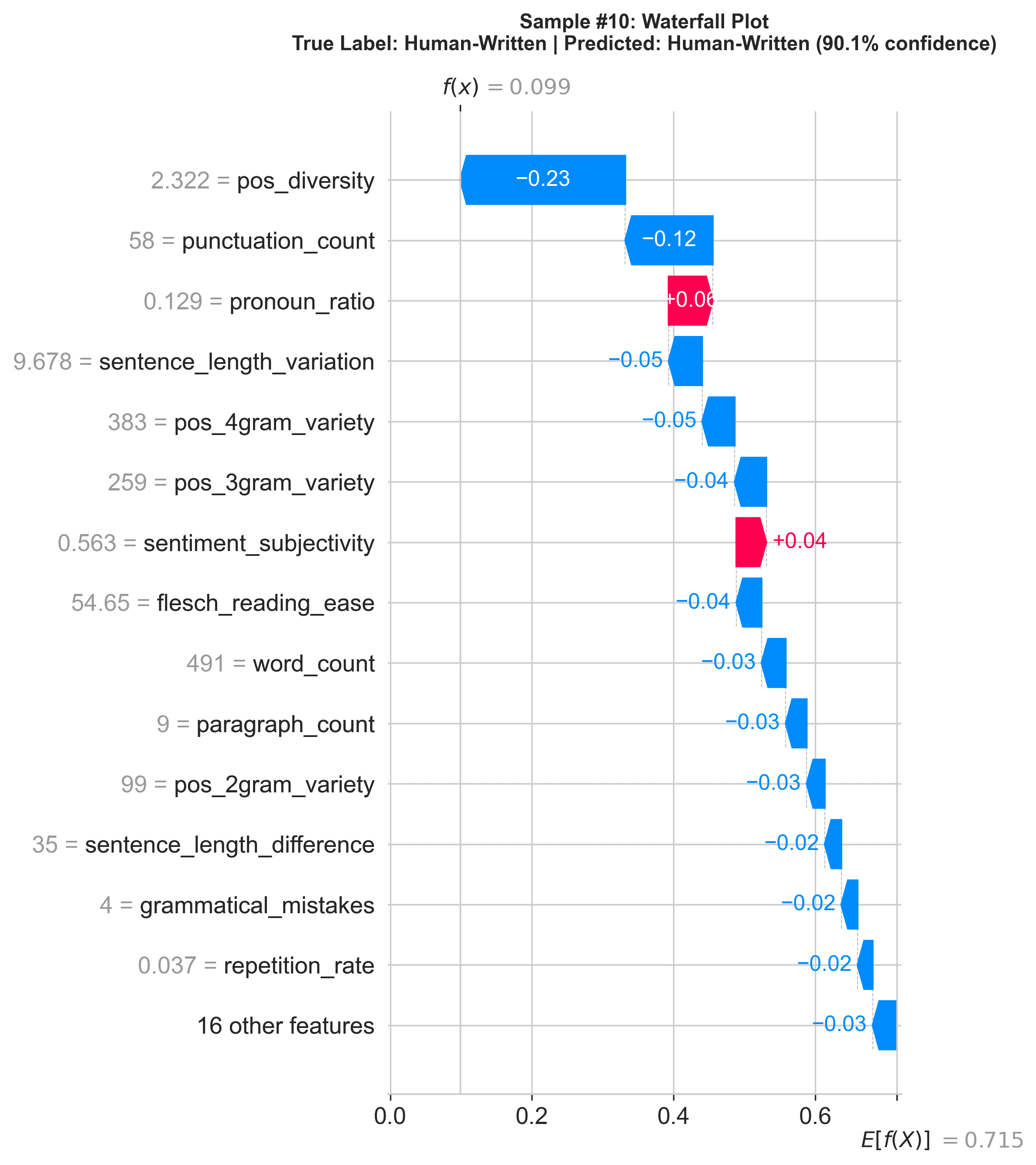}
        \caption{Human-written sample}
        \label{waterfall_human}
    \end{subfigure}\hfill
    \begin{subfigure}{0.48\textwidth}
        \centering
        \includegraphics[width=\linewidth]{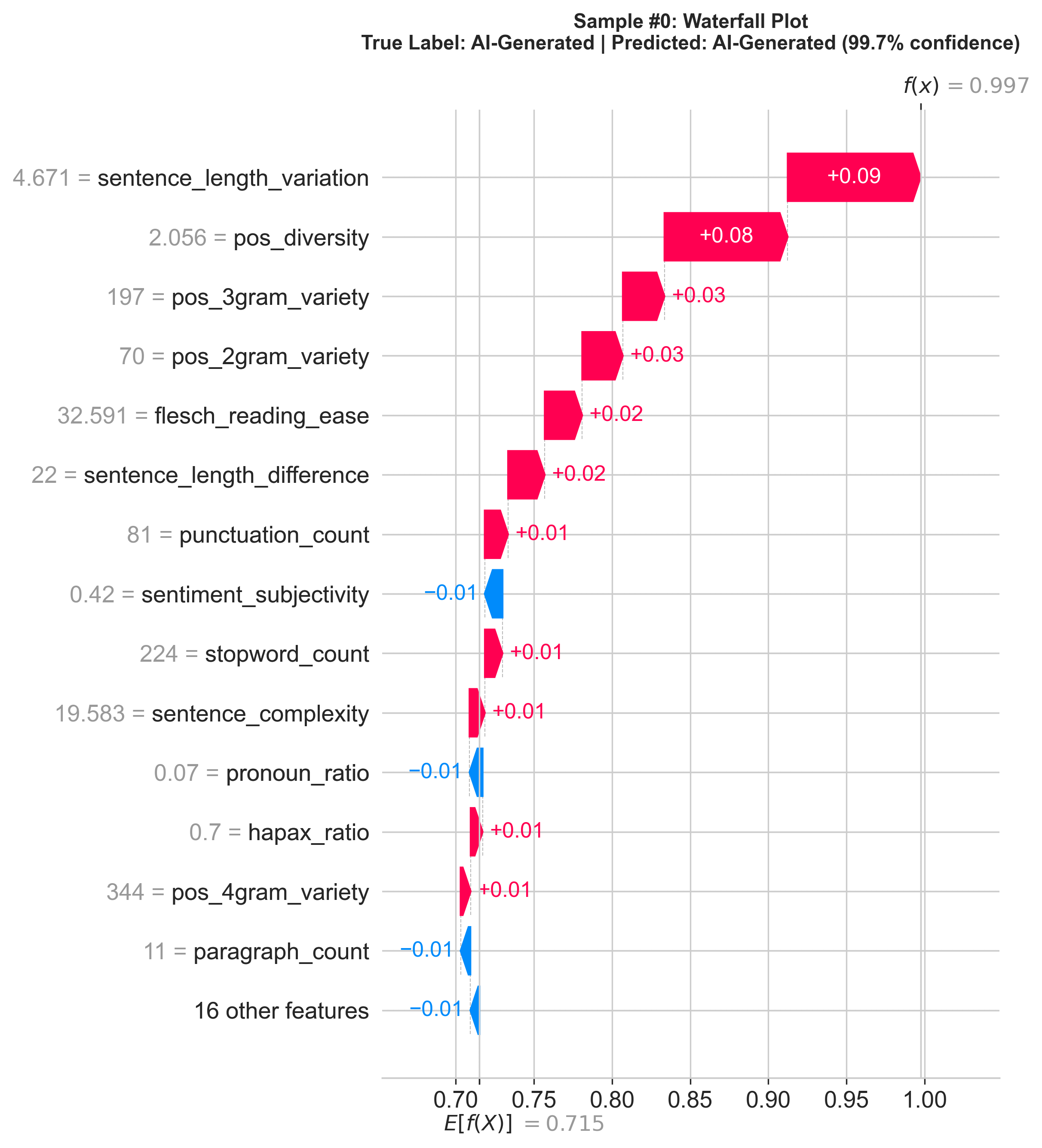}
        \caption{AI-generated sample}
        \label{waterfall_ai}
    \end{subfigure}
    \caption{SHAP waterfall plots illustrating feature contributions for human-written and AI-generated samples.}
    \label{fig:waterfall_plots}
\end{figure}

\begin{figure}[htbp]
    \centering
    \begin{subfigure}{0.5\textwidth}
        \centering
        \includegraphics[width=\linewidth]{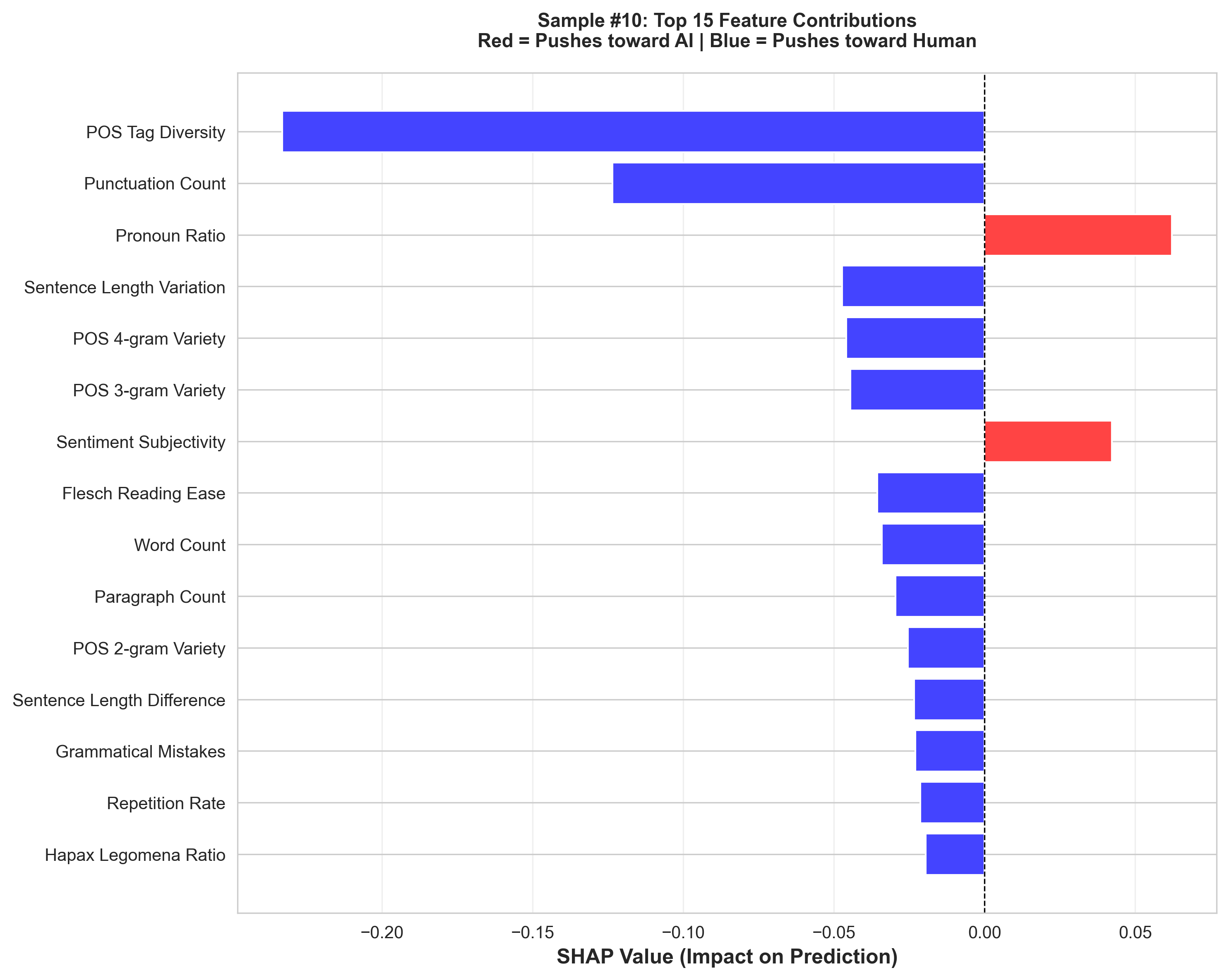}
        \caption{Human-written sample}
        \label{bar_human}
    \end{subfigure}\hfill
    \begin{subfigure}{0.48\textwidth}
        \centering
        \includegraphics[width=\linewidth]{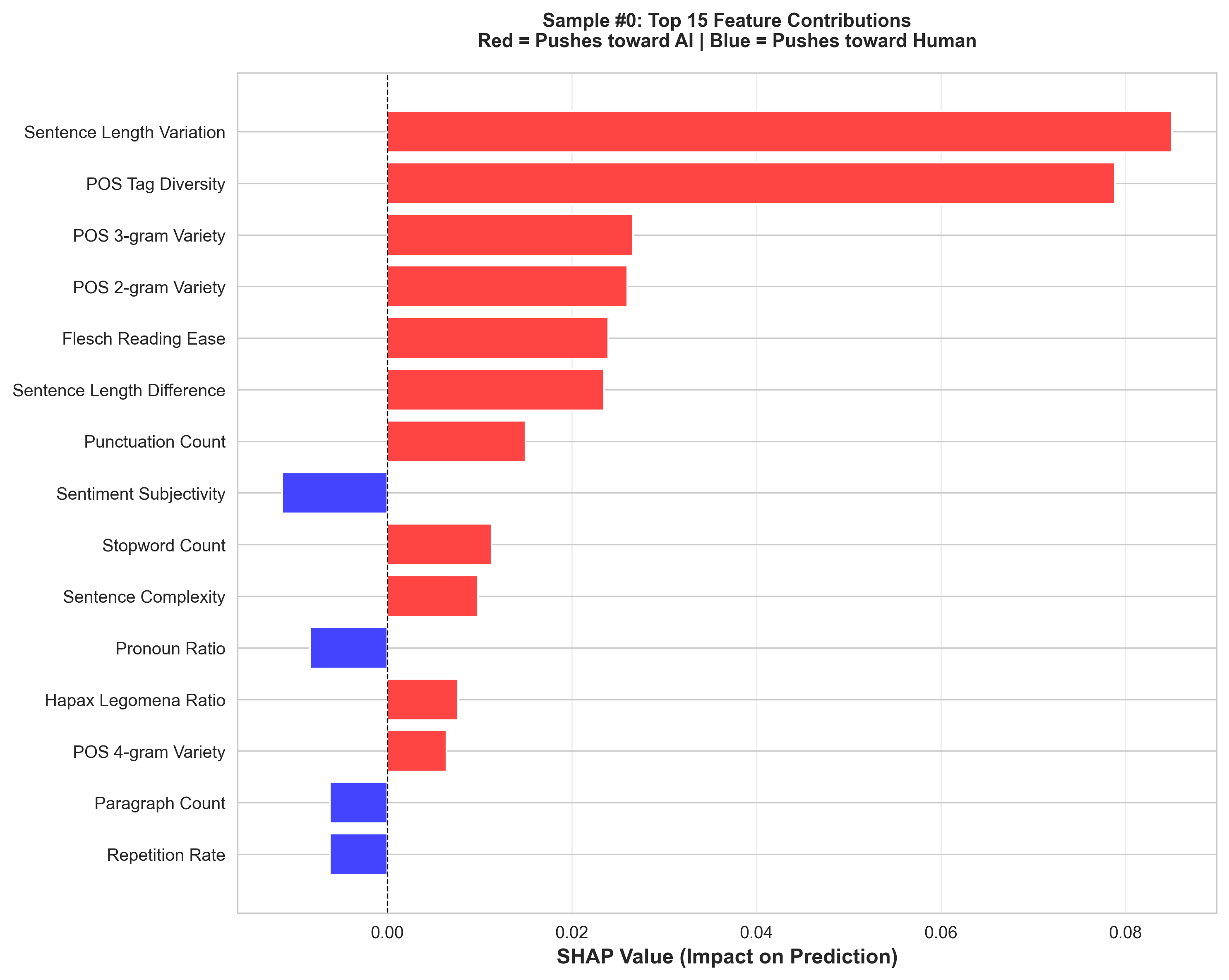}
        \caption{AI-generated sample}
        \label{bar_ai}
    \end{subfigure}
    \caption{SHAP bar plots showing average feature impact for human-written and AI-generated samples.}
    \label{fig:bar_plots}
\end{figure}

Instance-level explanations further illustrate this behaviour. SHAP waterfall and bar plots (Figures~\ref{fig:waterfall_plots} and~\ref{fig:bar_plots}) reveal that individual predictions depend on combinations of stylistic features such as sentence variation, punctuation usage, and syntactic diversity. These signals vary across domains, explaining why models succeed on benchmarks yet fail under domain shift. With such analysis, we conclude that the reason of failure in generalisation is that the algorithms are learning dataset-specific patterns rather than machine authorship.

\subsection{In Depth Error Analysis Using SHAP}
Extending the global level SHAP analysis, we analyse the test results of the highest performing model, i.e., the Random Forest model, trained on the COLING test set, to further analyse why the algorithm is not able to learn the genuine machine authorship and is only reflecting data-specific patterns. Paragraph Count and GZIP Compression Ratio being the top 2 important features from the global SHAP plot in the Fig \ref{fig:shap_coling}, we calculate the mean and median of these features in all categories (TP, FP, FN, and TN) along with word count, which is reported in the Table \ref{tab:text_class_stats}.

\begin{table}[htbp]
\centering
\caption{Descriptive statistics for text classification categories, including mean and median values for word count, paragraph count, and GZIP compression ratio. This table highlights structural differences between true positives (TP), true negatives (TN), false positives (FP), and false negatives (FN).}
\label{tab:text_class_stats}
\begin{tabular}{l c c c c c c c}
\hline
\textbf{Type} & \textbf{Count} & \makecell{\textbf{Words} \\ \textbf{(Mean)}} & \makecell{\textbf{Words} \\ \textbf{(Median)}} & \makecell{\textbf{Paragraph} \\ \textbf{(Mean)}} & \makecell{\textbf{Paragraph} \\ \textbf{(Median)}} & \makecell{\textbf{GZIP} \\ \textbf{(Mean)}} & \makecell{\textbf{GZIP} \\ \textbf{(Median)}} \\
\hline
FN & 139  & 177.48 & 183 & 13.24 & 3  & 0.724 & 0.557 \\
FP & 2753 & 221.48 & 171 & 4.13  & 1  & 0.574 & 0.548 \\
TN & 967  & 421.03 & 212 & 24.69 & 17 & 0.628 & 0.518 \\
TP & 6141 & 242.89 & 198 & 5.36  & 1  & 0.535 & 0.510 \\
\hline
\end{tabular}
\end{table}

The SHAP-based error analysis, combined with aggregate statistics across all 10,000 test samples (2,753 false positives, 139 false negatives, 6,141 true positives, and 967 true negatives), reveals that detector failures follow systematic, interpretable patterns rather than occurring at random. We identify three major root causes, each supported by convergent evidence from SHAP attribution analysis and population-level distributional statistics.

\textbf{Failure Reason I: Paragraph Count—the globally most important feature—is a formatting artefact, not a linguistic one.}

Corizzo et al. identified paragraph count as a key text-level feature, highlighting its role in capturing structural differences between human and AI-generated essays \cite{corizzo2023oneclass}. In our in-depth analysis, paragraph count shows the largest difference in SHAP values between false positives and false negatives ($\Delta$SHAP = +0.109), and it consistently emerges as the most influential feature in individual SHAP waterfall plots, with contributions ranging from 0.091 to 0.250.

Examining the aggregate statistics helps explain this behaviour. The model appears to have learned a strong association between single-paragraph texts and AI-generated content. For instance, true positives have a median paragraph count of 1, whereas true negatives have a median of 17—a substantial gap that the classifier uses as a key decision signal.

Both types of errors, however, deviate from this learned pattern. False positives also have a median paragraph count of 1, making these human-written texts structurally very similar to correctly identified AI texts based on this feature alone. In contrast, false negatives exhibit a different trend: their mean paragraph count is 13.2, which is more than double that of true positives (5.4). This suggests that AI-generated texts adopting a multi-paragraph structure begin to resemble human-written texts, increasing the likelihood of misclassification.

% Further, to understand the failure modes of the detector, we conducted a systematic SHAP-based analysis of misclassified samples. Five false positive (FP) cases—human-written texts incorrectly classified as AI-generated—and five false negative (FN) cases—AI-generated texts incorrectly classified as human-written—were selected from the test set, spanning a range of classifier confidence levels.

\textbf{Failure Reason II: GZIP Compression Ratio captures textual regularity, but this signal is domain-confounded.}

\noindent
Compression ratio emerges as the second most influential factor contributing to misclassification ($\Delta$SHAP = +0.097). However, a closer examination of the aggregate statistics suggests that this feature is inherently unstable as a reliable discriminator. The model appears to associate lower compression ratios with AI-generated text (true positive mean = 0.535) and higher ratios with human-written text (true negative mean = 0.628).

\medskip

\noindent
Notably, false negatives exhibit the highest mean compression ratio among all categories (0.724), which is approximately 35\% higher than the true positive mean. This pattern indicates that AI-generated texts which evade detection tend to display unusually high lexical diversity or structural variability, pushing their compression ratios into a range more typical of human-authored content.

\medskip

\noindent
These observations suggest that compression ratio does not capture a stable generative signature of AI text. Instead, it reflects a more complex interaction between text length and domain-specific characteristics, limiting its robustness as a standalone feature for classification.

\textbf{Failure Reason III: Feature Degradation Below a Minimum Viable Text Length.}

\noindent
Text length also plays a significant role in misclassification patterns. Both error categories tend to be substantially shorter than their correctly classified counterparts. Specifically, false negatives have an average length of 177 words compared to 243 words for true positives (a reduction of approximately 27\%), while false positives average 221 words compared to 421 words for true negatives (a reduction of 47\%).

\medskip

\noindent
A closer examination of the distribution reveals that the modal values provide additional insight. The most frequent lengths are 14 words for false negatives and 34 words for false positives, suggesting that many misclassifications occur in very short texts. Such texts often do not provide sufficient linguistic or structural information for reliable feature extraction.

\medskip

\noindent
In this context, short human-written texts may lack sufficient signal to be confidently identified as human, while short AI-generated texts may evade detection for the same reason. However, due to an apparent prior bias toward predicting the AI class, misclassification of short human texts (false positives) becomes more prevalent.

\medskip

\noindent
These findings highlight an important practical implication: enforcing a minimum text length threshold could improve classification reliability by ensuring that sufficient information is available for robust feature extraction.

\subsection{Discussion}

Taken together, the experiments reveal a consistent pattern. Models achieve near state-of-the-art accuracy on benchmark datasets, yet their performance degrades substantially when applied to new domains or generators. Explanations show that the most influential features differ across datasets, and the in-depth error analysis shows the major reason for failure is that the features that are most discriminative on in-domain data are also the features most susceptible to domain shift, formatting variation, and text-length effects. 

These findings support the dataset artefact hypothesis: many detectors distinguish particular corpora rather than reliably identifying machine authorship. The detector performs well not because it has learned what AI-generated language is, but because it has learned what the AI-generated texts in the training data look like. Benchmark accuracy, therefore, overestimates real-world detection capability.

We observe a systematic trade-off. Tree-based models optimise in-domain performance but overfit dataset-specific patterns. Linear models generalise more reliably but achieve lower peak accuracy. Ensemble methods improve robustness, yet they do not fully resolve the underlying validity problem.

Explainability plays a crucial role beyond transparency. Rather than merely justifying predictions, SHAP explanations function as a diagnostic tool, revealing the linguistic evidence used by the models and exposing why generalisation fails.

Our results imply that AI-text detectors should not be treated as forensic proof of misconduct. Even when a detector achieves very high in-domain benchmark scores, non-trivial false positives can occur, and performance can degrade sharply under domain or generator shift. In practice, a detector output should be interpreted as a probabilistic signal that must be contextualised with the assessment design, writing process evidence, and human review. For high-stakes decisions, institutions should require (i) cross-domain evaluation evidence relevant to the deployment setting, (ii) transparent instance-level explanations that indicate what linguistic cues influenced the decision, and (iii) policies that explicitly prohibit automated punitive action based solely on detector scores.

This SHAP-based error analysis highlights several implications for improving AI-generated text detection systems. The short-text failure mode indicates that detectors should apply minimum text-length thresholds or length-aware feature normalisation, as linguistic features become unreliable for very short inputs. Additionally, false positives in technical and domain-specific texts reveal that constrained vocabularies can resemble AI-generated patterns. Also, highly structured AI outputs demonstrate that formatting cues can mislead detectors, emphasising the importance of shifting toward content-level linguistic and semantic features rather than surface-level structural indicators. We believe these implications can lead to more robust detection.

% We hope these implications will lead to building a robust AI detector. 

Several limitations should be acknowledged. The research is based on English text and may be less stable for short documents where linguistic features are sparse. In addition, as language models evolve, stylistic distributions may change, requiring periodic re-evaluation of detection systems. Finally, the detector should be interpreted as probabilistic evidence supporting human judgment rather than an automated mechanism for punitive decisions in educational contexts.

Overall, the results suggest that reliable deployment of AI-text detectors requires generalisation-aware evaluation and interpretable evidence, not benchmark accuracy alone.

\FloatBarrier

\section{Conclusion and Future Work}

This paper investigated not only how accurately AI-generated text can be detected, but also what current detection systems actually learn. We introduced an interpretable detection framework based on linguistic features, classical machine learning, and instance-level explainable AI. Across multiple benchmark datasets, the proposed approach achieved competitive performance, reaching an F1 score of 97.34 on the PAN CLEF benchmark while providing transparent explanations for individual predictions.

More importantly, the experiments revealed a consistent pattern across evaluation settings. Models achieved near state-of-the-art accuracy when trained and tested on the same corpus, yet performance degraded substantially under domain and generator shift. Explanations further showed that the most influential features differed markedly between datasets. Together, these findings indicate that many detectors rely on dataset-specific stylistic cues rather than stable indicators of machine authorship. Consequently, benchmark accuracy alone is not a reliable measure of real-world detection capability.

The results, therefore, highlight a fundamental trade-off between optimisation for benchmark performance and robustness to new data. Tree-based models achieved the highest in-domain accuracy but overfit corpus-specific patterns, whereas linear models generalised more consistently. Ensemble methods improved robustness but did not eliminate the underlying dependence on dataset characteristics. Explainable AI proved essential in diagnosing this behaviour: rather than serving only as a transparency mechanism, instance-level explanations provided evidence about the linguistic signals actually used by the models.

From a practical perspective, the framework offers a transparent alternative to black-box detectors. The released open-source implementation allows predictions to be accompanied by interpretable evidence, supporting informed human judgement in educational and other high-stakes settings. The detector should therefore be viewed as an assistive analytic tool rather than an automated decision system.

% Several limitations should be acknowledged. The present study focuses on English-language text and document-level linguistic features, which may be less stable for very short inputs. In addition, as generative models evolve, stylistic distributions may change, requiring periodic re-evaluation of detection approaches. The results also suggest that no single detector is likely to remain reliable across all future models and domains.

Future work should therefore move beyond maximising benchmark accuracy toward evaluating detection validity. Promising directions include multilingual analysis, shorter text detection, and adaptive models that explicitly account for domain variation. Further research is also needed on evaluation protocols, particularly the development of cross-domain benchmarks and longitudinal testing against newly released language models. Finally, integrating detection with pedagogical strategies and human review processes may provide a more reliable approach than relying on automated classification alone.

Overall, this work demonstrates that interpretable analysis is essential for understanding and responsibly deploying AI-generated text detection. Reliable use of such systems requires not only accurate predictions, but evidence that the predictions reflect genuine authorship signals rather than dataset artefacts.

\section{Acknowledgement}
This publication has emanated from research [conducted with the financial support of/supported in part by a grant from] Science Foundation Ireland under Grant number 18/CRT/6183. For the purpose of Open Access, the author has applied a CC BY public copyright licence to any Author Accepted Manuscript version arising from this submission.

\bibliographystyle{unsrt}  
\bibliography{references}

\end{document}